\documentclass{article}

\usepackage{microtype}
\usepackage{graphicx}
\usepackage{subfigure}
\usepackage{booktabs} 

\usepackage{hyperref}

\usepackage[accepted]{icml2024}

\usepackage{amsmath}
\usepackage{amssymb}
\usepackage{mathtools}
\usepackage{amsthm}

\usepackage[capitalize,noabbrev]{cleveref}

\theoremstyle{plain}
\newtheorem{theorem}{Theorem}[section]
\newtheorem{proposition}[theorem]{Proposition}

\theoremstyle{definition}

\newtheorem{problem}[theorem]{Problem}
\theoremstyle{remark}

\usepackage[textsize=tiny]{todonotes}

\usepackage{url}            
\usepackage{amscd}
\usepackage{mathrsfs}
\usepackage{algorithm,algorithmic}
\renewcommand{\algorithmicrequire}{\textbf{Input:}}
\renewcommand{\algorithmicensure}{\textbf{Output:}}

\usepackage{tikz}
\usepackage{bm}
\usepackage[export]{adjustbox}
\usepackage{multirow}
\newcommand{\set}[1]{\{ #1 \}}
\newcommand{\ind}[1]{\mathbb{I}\left[ #1 \right]}

\DeclareMathOperator*{\argmin}{\operatorname{arg\, min}}
\DeclareMathOperator*{\argmax}{\operatorname{arg\, max}}

\renewcommand{\epsilon}{\varepsilon}
\newcommand{\bx}{\bm{x}}
\newcommand{\ba}{\bm{a}}
\newcommand{\haty}{\hat{y}}
\newcommand{\hatyl}{\hat{y}_\mathrm{L}}
\newcommand{\hatyr}{\hat{y}_\mathrm{R}}
\newcommand{\calA}{\mathcal{A}}

\crefname{algorithm}{Algorithm}{Algorithms}
\crefname{table}{Table}{Tables}
\crefname{figure}{Figure}{Figures}
\crefname{section}{Section}{Sections}
\crefname{appendix}{Appendix}{Appendices}
\crefname{problem}{Problem}{Problems}
\crefname{theorem}{Theorem}{Theorems}
\crefname{proposition}{Proposition}{Propositions}
\crefname{lemma}{Lemma}{Lemmas}
\crefname{remark}{Remark}{Remarks}

\icmltitlerunning{Learning Decision Trees and Forests with Algorithmic Recourse}

\begin{document}

\twocolumn[
\icmltitle{Learning Decision Trees and Forests with Algorithmic Recourse}




\begin{icmlauthorlist}
\icmlauthor{Kentaro Kanamori}{fj}
\icmlauthor{Takuya Takagi}{fj}
\icmlauthor{Ken Kobayashi}{tit}
\icmlauthor{Yuichi Ike}{ku}
\end{icmlauthorlist}

\icmlaffiliation{fj}{Fujitsu Limited, Japan}
\icmlaffiliation{tit}{Tokyo Institute of Technology, Japan}
\icmlaffiliation{ku}{Kyushu University, Japan}

\icmlcorrespondingauthor{Kentaro Kanamori}{k.kanamori@fujitsu.com}

\icmlkeywords{algorithmic recourse, tree-based model, counterfactual explanation}

\vskip 0.3in
]



\printAffiliationsAndNotice{}  

\begin{abstract}
This paper proposes a new algorithm for learning accurate tree-based models while ensuring the existence of recourse actions. 
Algorithmic Recourse (AR) aims to provide a recourse action for altering the undesired prediction result given by a model. 
Typical AR methods provide a reasonable action by solving an optimization task of minimizing the required effort among executable actions. 
In practice, however, such actions do not always exist for models optimized only for predictive performance. 
To alleviate this issue, we formulate the task of learning an accurate classification tree under the constraint of ensuring the existence of reasonable actions for as many instances as possible. 
Then, we propose an efficient top-down greedy algorithm by leveraging the adversarial training techniques. 
We also show that our proposed algorithm can be applied to the random forest, which is known as a popular framework for learning tree ensembles. 
Experimental results demonstrated that our method successfully provided reasonable actions to more instances than the baselines without significantly degrading accuracy and computational efficiency. 
\end{abstract}

\section{Introduction}
\begin{figure*}
    \centering
    \subfigure[Instance $\bx$]{
        \adjustbox{valign=b}{
            \begin{tabular}{lc}
            \toprule
                \textbf{Features} & \textbf{Values} \\
            \midrule
               Income & \$70K \\
               Purpose & NewCar \\
               Education & College \\
               \#ExistingLoans & 2 \\
            \bottomrule
            \end{tabular}
            \label{fig:intro:demo:instance}    
        }
    }
    \hfill
    \subfigure[Classification tree $h_0$]{
        \includegraphics[width=0.3\textwidth,valign=b]{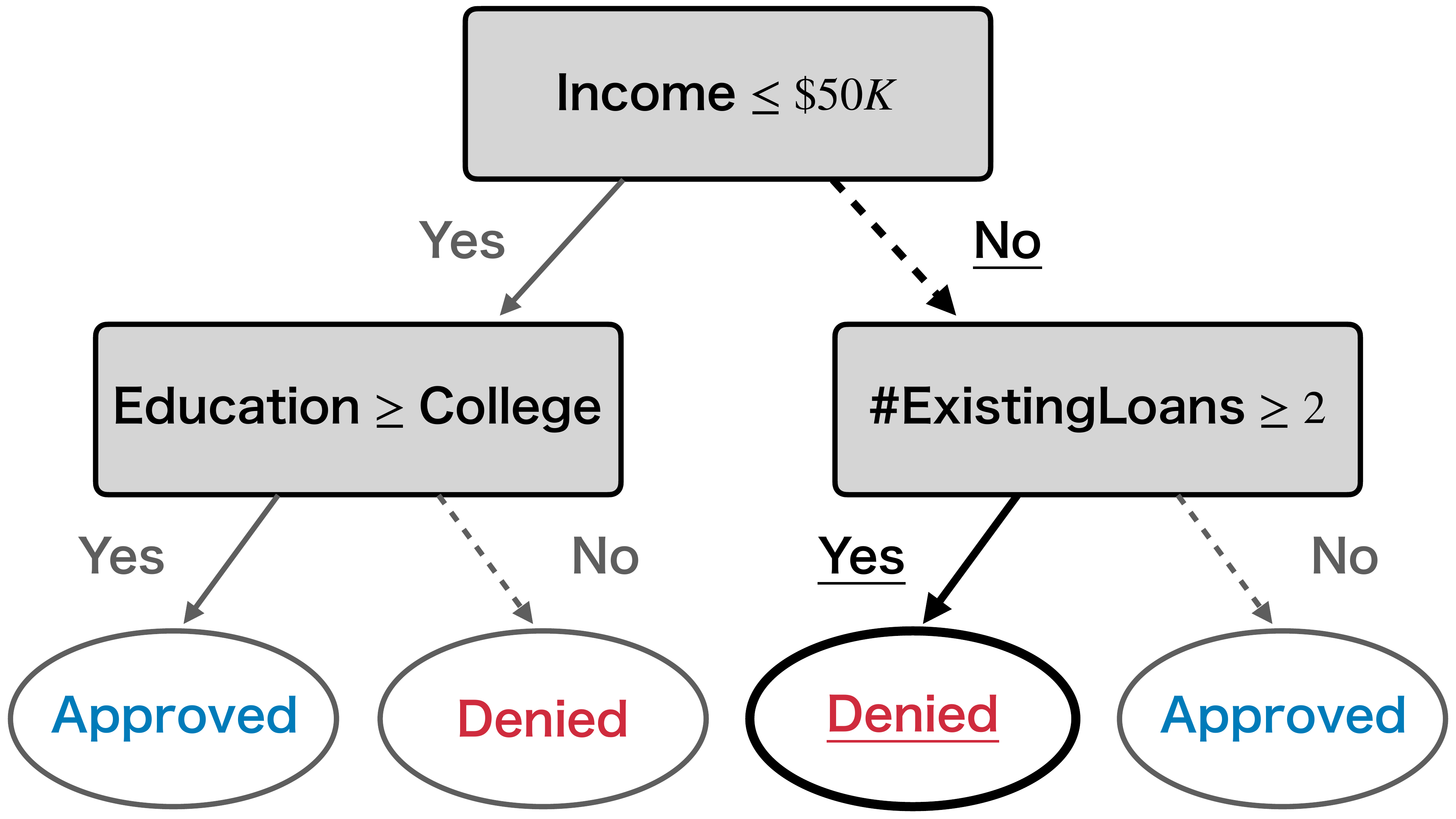}
        \label{fig:intro:demo:tree1}    
    }
    \hfill
    \subfigure[Classification tree $h_1$]{
        \includegraphics[width=0.3\textwidth,valign=b]{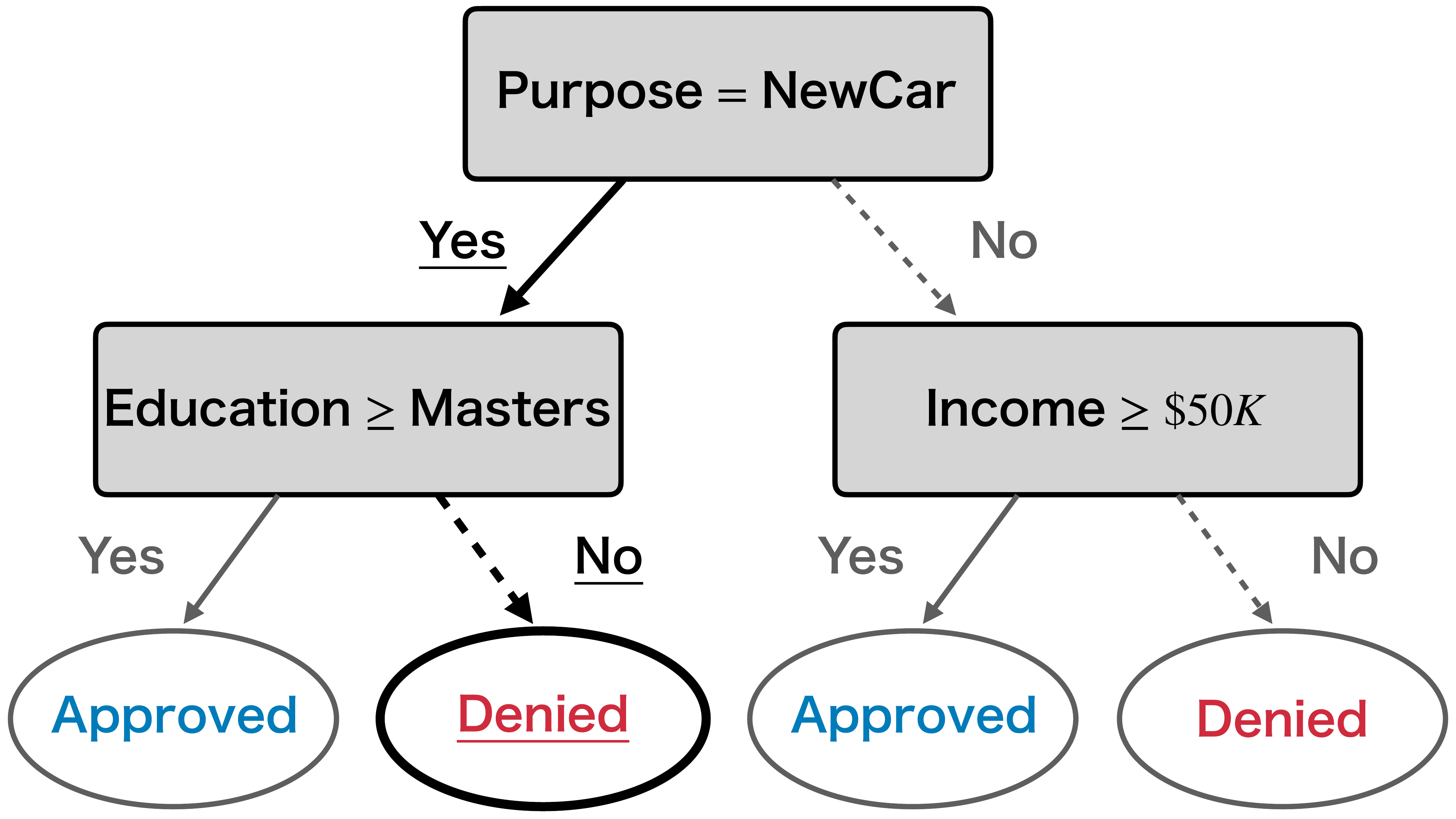}
        \label{fig:intro:demo:tree2}    
    }
    \caption{
        Examples of an input instance $\bx$ and classification trees $h_0, h_1$. 
        To get the loan approved from $h_0$, the instance $\bx$ needs to just reduce ``\#ExistingLoans." 
        In contrast, for the case of $h_1$, $\bx$ needs to change ``Purpose" or ``Education," which are not executable easily. 
    }
    \label{fig:intro:demo}
\end{figure*}

Algorithmic decision-making with machine learning models has been applied to various tasks in the real world, such as loan approvals. 
In such critical tasks, the predictions made by a model might have a significant impact on individual users~\cite{Rudin:NMI2019}. 
Consequently, decision-makers need to explain how individuals should act to alter the undesired decisions~\cite{Miller:AI2019,Wachter:HJLT2018}. 
\emph{Algorithmic Recourse~(AR)} aims to provide such information~\cite{Ustun:FAT*2019}. 
For a classifier $h \colon \mathcal{X} \to \mathcal{Y}$, a desired class $y^\ast \in \mathcal{Y}$, and an instance $\bx \in \mathcal{X}$ such that $h(\bx) \not= y^\ast$, AR provides a perturbation $\ba$ that flips the prediction result into the desired class, i.e., $h(\bx + \ba) = y^\ast$, with a minimum effort measured by some cost function $c$. 
The user can regard the perturbation $\ba$ as a \emph{recourse action} for obtaining the desired outcome $y^\ast$ from the classifier $h$~\cite{Karimi:ACMCS2022}. 

To provide affected individuals with recourse certainly, we need to guarantee the existence of actions $\ba$ that are executable with small costs. 
In general, however, such reasonable actions do not always exist~\cite{Ross:NIPS2021}. 
\cref{fig:intro:demo} demonstrates this observation on a synthetic loan approval task. 
Consider a situation where a user $\bx$ in \cref{fig:intro:demo:instance} is denied one's loan application by classification trees $h_0$ and $h_1$ in \cref{fig:intro:demo:tree1,fig:intro:demo:tree2}. 
By changing ``\#ExistingLoans," the user can get the loan approved by $h_0$. 
In contrast, for the case of $h_1$, the user must change ``Purpose" or ``Education," which are inappropriate recourse actions because changing these features is difficult in practice.  
It suggests that the user $\bx$ cannot get one's loan approved as long as $h_1$ is deployed, and that $h_0$ is more desirable as it can ensure a reasonable action for the user $\bx$~\cite{Sullivan:AIES2022,Venkatasubramanian:FAT*2020}. 

In this paper, we aim to learn a classifier $h$ that ensures the existence of executable actions $\ba$ for input instances $\bx$ with high probability while keeping accuracy. 
Specifically, we focus on the \emph{tree-based models}, such as \emph{decision trees}~\cite{Breiman:1984:CART} and \emph{random forests}~\cite{Breiman:ML2001}. 
Recently, \citet{Ross:NIPS2021} proposed a gradient-based method for learning differentiable classifiers such as deep neural networks while guaranteeing the existence of actions. 
However, their method cannot be directly applied to the tree-based models because these models are not differentiable. 
Tree-based models are known to perform well on tabular datasets~\cite{Grinsztajn:NIPS2022}, which often appear in the areas where AR is required (e.g., finance and justice)~\cite{Verma:arxiv2020}. 
Therefore, we need a new learning algorithm designed for tree-based models while ensuring the existence of recourse actions.

\subsection{Our Contributions}
In this paper, we propose \emph{Recourse-Aware Classification Tree~(RACT)}, a new framework for learning tree-based classifiers that make accurate predictions and guarantee recourse actions. 
Our contributions are summarized as follows:
\begin{itemize}
    \item 
    We introduce the task of learning a classification tree under the constraint on the ratio of instances having at least one valid action. 
    Then, we propose a specialized top-down greedy algorithm by leveraging the adversarial training techniques~\cite{Guo:ICML2022}. 
    Our algorithm is simple yet efficient, as we show that its computational complexity is equivalent to that of the standard learning algorithm for classification trees, such as CART~\cite{Breiman:1984:CART}. 
    \item 
    We introduce a post-processing task of modifying a learned tree so as to satisfy the constraint on recourse guarantee, and prove that the task can be efficiently solved with an approximation guarantee by reducing it to the minimum set cover problem~\cite{Kearns:1990:computational}. 
    We also show that our learning algorithm can be easily extended to the random forest~\cite{Breiman:ML2001}, which is a popular framework for learning tree ensembles. 
    \item 
    Our numerical experiments on real datasets demonstrated that RACT successfully provided reasonable recourse actions to more individuals than the baselines while keeping comparable predictive accuracy and computational efficiency.
    We also confirmed that we could control the trade-off between keeping accuracy and guaranteeing recourse by tuning the hyperparameter of RACT.
\end{itemize}

\subsection{Related Work}
\emph{Algorithmic Recourse (AR)}~\cite{Ustun:FAT*2019}, also referred to as \emph{Counterfactual Explanation}~\cite{Wachter:HJLT2018}, has attracted increasing attention in recent years~\cite{Verma:arxiv2020,Karimi:ACMCS2022}. 
While almost all of the previous papers focused on extracting actions from a learned machine learning model~\cite{Cui:KDD2015,Tolomei:KDD2017,Pawelczyk:WWW2020,Kanamori:IJCAI2020,Kanamori:AAAI2021,Parmentier:ICML2021,Lucic:AAAI2022,Carreira:AAAI2023}, there exist few studies on learning models while ensuring the existence of recourse actions~\cite{Ross:NIPS2021,Dominguez-olmedo:ICML2022}. 
However, because the existing methods are based on gradient descent, we cannot directly apply them to learning tree-based models. 

Learning \emph{tree-based models}, such as decision trees~\cite{Breiman:1984:CART}, random forests~\cite{Breiman:ML2001}, and gradient boosted trees~\cite{Friedman:AS2000,Chen:KDD2016,Ke:NIPS2017}, has been widely studied due to their interpretability, accuracy, and efficiency~\cite{Hu:NIPS2019,Grinsztajn:NIPS2022}. 
There also exist several methods for learning tree-based models under some constraints, such as fairness~\cite{Kamiran:ICDM2010} and stability~\cite{Hara:ICLR2023}. 
Our method is most related to the algorithms for learning robust decision trees~\cite{Chen:ICML19,Vos:ICML2021,Vos:ECMLPKDD2022,Guo:ICML2022}. 
To learn a robust decision tree, these methods determine the split condition in each node by optimizing the worst-case information gain under adversarial perturbations. 
In contrast, we aim to ensure the existence of executable actions, i.e., some kind of perturbations, for obtaining a desired outcome.  
Thus, it is not trivial to apply these methods to our task. 


\section{Problem Statement}
For a positive integer $n \in \mathbb{N}$, we write $[n] \coloneqq \set{1, \dots, n}$.
As with the previous studies~\cite{Ross:NIPS2021}, we consider a binary classification task between undesired and desired classes.  
Let $\mathcal{X} \subseteq \mathbb{R}^{D}$ and $\mathcal{Y} = \set{\pm 1}$ be input and output domains, respectively. 
We assume that all the categorical features are encoded by one-hot encoding. 
We call a vector $\bx = (x_1, \dots, x_D) \in \mathcal{X}$ an \emph{instance}, and a function $h \colon \mathcal{X} \to \mathcal{Y}$ a \emph{classifier}. 
We denote the $0$-$1$ loss function by $l_{01}(y, \haty) = \ind{y \not= \haty}$. 
Without loss of generality, we assume that $h(\bx)=+1$ is a desirable result (e.g., loan approval).

\subsection{Algorithmic Recourse}
For an instance $\bx \in \mathcal{X}$, we define an \emph{action} as a perturbation vector $\ba \in \mathbb{R}^{D}$ such that $\bx+\ba \in \mathcal{X}$. 
Let $\calA(\bx)$ be a set of feasible actions for $\bx$ such that $\bm{0} \in \calA(\bx)$ and $\calA(\bx) \subseteq \{ \ba \in \mathbb{R}^{D} \mid \bx + \ba \in \mathcal{X} \}$. 
For simplicity, we assume that we can write as $\calA(\bx) = [l_1, u_1] \times \dots \times [l_D, u_D]$ with lower and upper bounds $l_d, u_d \in \mathbb{R}$ for $d \in [D]$. 
For a classifier $h$, an action $\ba$ is \emph{valid} for $\bx$ if $\ba \in \calA(\bx)$ and $h(\bx + \ba)=+1$. 
For $\bx \in \mathcal{X}$ and $\ba \in \calA(\bx) $, a \emph{cost function} $c \colon \calA(\bx) \to \mathbb{R}_{\geq 0}$ measures the required effort of $\ba$ with respect to $\bx$. 
For a given classifier $h \colon \mathcal{X} \to \mathcal{Y}$ and an instance $\bx \in \mathcal{X}$, the aim of \emph{Algorithmic Recourse (AR)} is to find an action $\ba$ that is valid for $\bx$ with respect to $h$ and minimizes its cost $c(\ba \mid \bx)$. 
This task can be formulated as follows~\cite{Karimi:ACMCS2022}:
\begin{align}\label{eq:ce}
    \min_{\ba \in \calA(\bx)} \; c(\ba \mid \bx) \;\;\; \text{s.t.} \;\; h(\bx + \ba) = +1.
\end{align}

For the cost function $c$, we assume that it satisfies the following properties: 
(i)~$c(\bm{0} \mid \bx) = 0$;
(ii)~$c(\ba \mid \bx) = \max_{d \in [D]} c_d(a_d \mid x_d)$, where $c_d$ is the cost of the action $a_d$ for a feature $d$;
(iii)~$\forall \beta \geq 0: c_d(a_d \mid x_d) \leq c_d(a_d \cdot (1 + \beta) \mid x_d)$. 
Note that several major cost functions, including the weighted $\ell_\infty$-norm~\cite{Ross:NIPS2021} and max percentile shift~\cite{Ustun:FAT*2019}, satisfy the above properties. 

\subsection{Tree-Based Model}
A \emph{classification tree} $h \colon \mathcal{X} \to \mathcal{Y}$ is a classifier that consists of a set of if-then-else rules expressed as a binary tree structure~\cite{Breiman:1984:CART}.
For a given input $\bx \in \mathcal{X}$, it makes a prediction according to the predictive label $\haty$ of the leaf that $\bx$ reaches. 
The corresponding leaf is determined by traversing the tree from the root depending on whether the \emph{split condition} $x_d \leq b$ is true or not, where $(d, b) \in [D] \times \mathbb{R}$ is a pair of a feature and threshold of each internal node. 
\cref{fig:intro:demo} presents examples of classification trees. 

For a classification tree $h$, we denote the total number of its leaves by $I \in \mathbb{N}$.
Let $r_i \subseteq \mathcal{X}$ and $\haty_i \in \mathcal{Y}$ be the subspace and predictive label corresponding to a leaf $i \in [I]$. 
Each subspace $r_i$ is determined by the split conditions on the path from the root to the leaf $i$, and can be expressed as an axis-aligned rectangle $r_i = (l_{i, 1}, u_{i, 1}] \times \dots (l_{i, D}, u_{i, D}]$. 
The set of such subspaces $\set{r_1, \dots, r_I}$ gives a partition of the input space $\mathcal{X}$, and $h$ can be expressed as $h(\bx) = {\sum}_{i = 1}^{I} \haty_i \cdot \ind{\bx \in r_i}$~\cite{Freitas:EN2014,Guidotti:CSUR2018}. 

A \emph{tree ensemble} is a popular machine learning model for tabular datasets~\cite{Grinsztajn:NIPS2022}. 
It makes a prediction by combining the outputs of $T$ trees $h_1, \dots, h_T$. 
In this paper, we focus on \emph{random forests}~\cite{Breiman:ML2001} as a framework for learning tree ensembles. 
Each tree $h_t$ of a random forest classifier is trained on a bootstrap sample of a training sample, and a split condition $(d, b)$ in each node is determined among a random subset of the features $[D]$. 

\subsection{Problem Formulation}
The aim of this paper is to learn an accurate tree-based classifier while guaranteeing the existence of a valid action with a low cost for any instance. 
For that purpose, we need to learn a classifier $h$ under the constraint that, for any instance $\bx$ in a given sample, there exists an action $\ba \in \calA(\bx)$ such that $h(\bx + \ba) = +1$ and $c(\ba \mid \bx) \leq \varepsilon$ for some budget parameter $\epsilon > 0$. 
In practice, however, previous studies have empirically demonstrated that it is often difficult to guarantee valid actions for all instances without degrading accuracy~\cite{Levanon:ICML2021,Olckers:arxiv2023}. 
This means that imposing the constraint of ensuring valid actions for all instances is too strict to keep accuracy. 

To balance such a trade-off, we will relax the constraint on recourse. 
For a cost budget parameter $\epsilon$, let $\calA_{\epsilon}(\bx) \coloneqq \set{\ba \in \calA(\bx) \mid c(\ba \mid \bx) \leq \epsilon}$ be the set of feasible actions whose costs are lower than $\epsilon$. 
Given a sample $S = \set{(\bx_n, y_n)}_{n=1}^{N}$, our \emph{empirical recourse risk} $\hat{\Omega}_{\epsilon}$ is defined as
\begin{align*}
    \hat{\Omega}_{\epsilon}(h \mid S) \coloneqq \frac{1}{N} {\sum}_{n=1}^{N} l_\mathrm{rec}(\bx_n; h), 
\end{align*}
where $l_\mathrm{rec}(\bx; h) \coloneqq {\min}_{\ba \in \calA_{\epsilon}(\bx)} l_{01}(+1, h(\bx + \ba))$ is the \emph{recourse loss}, which corresponds to the relaxed version of the original hard constraint $\exists \ba \in \calA_{\epsilon}(\bx): h(\bx + \ba) = +1$. 
By definition, this risk is equivalent to the ratio of instances $\bx$ in the sample $S$ that do not have any action $\ba$ such that $h(\bx + \ba) = +1$ and $c(\ba \mid \bx) \leq \varepsilon$. 

Using the empirical recourse risk $\hat{\Omega}_{\epsilon}$, we formulate our learning task as follows. 

\begin{problem}\label{prob:tears}
    Given a sample $S = \set{(\bx_n, y_n)}_{n=1}^{N} \subseteq \mathcal{X} \times \mathcal{Y}$ and parameters $\epsilon, \delta > 0$, 
    find an optimal solution to the following optimization problem:
    \begin{align*}
    \begin{array}{clcl}
        \displaystyle \mathop{\text{\upshape minimize}}_{h \in \mathcal{H}} & \hat{R}(h \mid S) & \text{subject to} & \hat{\Omega}_{\epsilon}(h \mid S) \leq \delta,
    \end{array}
    \end{align*}
    where 
    $\mathcal{H}$ is the set of classification trees $h \colon \mathcal{X} \to \mathcal{Y}$ and 
    $\hat{R}(h \mid S) \coloneqq \frac{1}{N} \sum_{n=1}^{N} l_{01}(y_n, h(\bx_n))$ is the empirical risk. 
\end{problem}

By solving \cref{prob:tears}, we can obtain a classification tree $h^\ast$ that makes accurate predictions while guaranteeing the existence of valid actions for at least $100 \cdot (1 - \delta)$ \% instances. 
We are also expected to adjust the trade-off between the predictive accuracy and recourse guarantee by tuning $\delta$. 



\section{Learning Algorithm}
This section presents an efficient algorithm for \cref{prob:tears}. 
Even for the case without the constraint $\hat{\Omega}_{\epsilon}(h \mid S) \leq \delta$, efficiently finding an exact optimal solution to \cref{prob:tears} is computationally challenging~\cite{Hyafil:IPL1976,Hu:NIPS2019}. 
In addition, the existing gradient-based methods, such as \cite{Ross:NIPS2021}, cannot be directly applied to \cref{prob:tears} due to the combinatorial nature of tree-based models. 
To avoid these difficulties, we propose an efficient learning algorithm by extending the standard greedy approach like CART~\cite{Breiman:1984:CART}. 
Our framework, named \emph{recourse-aware classification tree~(RACT)}, consists of the following two steps: 
(i)~growing a classification tree $h$ by recursively determining the split condition of each internal node based on the empirical risk $\hat{R}$ and recourse risk $\hat{\Omega}_\epsilon$; 
(ii)~modifying the predictive label of each leaf in the learned tree $h$ so as to satisfy the constraint $\hat{\Omega}_{\epsilon}(h \mid S) \leq \delta$. 
We also show that our algorithm can be applied to the random forest~\cite{Breiman:ML2001}, which is one of the most popular frameworks for learning tree ensembles. 
Note that our learning algorithm does not require a specific oracle to obtain an action (i.e., solve the problem \eqref{eq:ce} for each instance $\bx$) because our algorithm itself does not depend on the construction process of actions, and thus, we can employ any existing AR method for tree-based models (e.g., \cite{Cui:KDD2015,Tolomei:KDD2017,Kanamori:IJCAI2020}).

\subsection{Top-Down Greedy Splitting}
We first propose an efficient learning algorithm, inspired by the well-known top-down greedy approach to learning classification trees. 
For each node in a current tree $h$, it determines a split condition $(d, b)$ that minimizes some impurity criterion~\cite{Breiman:1984:CART,Rokach:TSMC2005}. 
Following this approach, we first formulate the task of determining a split condition designed for our learning problem, and propose an efficient algorithm for solving it. 

We consider adding a new split condition $(d, b)$ to a leaf $i \in [I]$ of a current tree $h$ with $I$ leaves. 
Such a new tree $h'$ can be expressed as follows: 
\begin{align*}
    h'(\bx) = \ind{\bx \in r_i} \cdot h(\bx; d, b, \hatyl, \hatyr) + \ind{\bx \not\in r_i} \cdot h(\bx), 
\end{align*}
where $h(\bx; d, b, \hatyl, \hatyr) = \hatyl \cdot \ind{\bx_d \leq b} + \hatyr \cdot \ind{\bx_d > b}$ is a decision stump with the split condition $(d, b)$ and predictive labels $\hatyl, \hatyr \in \mathcal{Y}$ corresponding to the left and right children of the node $i$, respectively. 
Our task can be formulated as finding a best split condition $(d, b)$ and predictive labels $\hatyl, \hatyr$. 
We assume that we are given a set $B_d = \set{b_{d, 1}, \dots, b_{d, J_d}}$ of thresholds for each feature $d \in [D]$ in advance. 
As with the previous studies, we also assume $b_{d, 1} < \dots < b_{d, J_d}$ and $J_d = \mathcal{O}(N)$. 
To relax the constraint $\hat{\Omega}_{\epsilon}(h \mid S) \leq \delta$ of \cref{prob:tears}, we define a penalized objective function as 
\begin{align*}
    \Phi_\lambda(d, b, \hatyl, \hatyr) \coloneqq \hat{R}(h' \mid S) + \lambda \cdot \hat{\Omega}_\epsilon(h' \mid S),     
\end{align*}
where $\lambda \geq 0$ is a hyper-parameter that balances the trade-off between the empirical risk $\hat{R}$ and recourse risk $\hat{\Omega}_\epsilon$. 
Then, our task is formulated as the following problem:
\begin{align}\label{eq:split}
    \begin{split}
        \min_{d \in [D], b \in B_d} \min_{\hatyl, \hatyr \in \mathcal{Y}} &\Phi_\lambda(d, b, \hatyl, \hatyr).
    \end{split}
\end{align}
When $\lambda = 0$, the problem \eqref{eq:split} is roughly equivalent to the standard learning problem for a classification tree, and can be solved in $\mathcal{O}(D \cdot N)$ if we have a permutation $\sigma_d \colon [N] \to [N]$ such that $x_{\sigma_d(1), d} \leq \dots \leq x_{\sigma_d(N), d}$ for each $d \in [D]$ in advance~\cite{Rokach:TSMC2005}. 
In contrast, if $\lambda > 0$, efficiently solving the problem \eqref{eq:split} is not trivial. 
In the following, we show that the problem \eqref{eq:split} with $\lambda>0$ can be solved in $\mathcal{O}(D \cdot N)$ as well by leveraging the adversarial training techniques for classification trees~\cite{Guo:ICML2022}. 

\begin{algorithm}[t]
    \caption{Algorithm for the problem \eqref{eq:split}. }
    \scriptsize
    \begin{algorithmic}[1]
        \REQUIRE{
            a sample $S = \set{ (\bx_n, y_n) }_{n=1}^{N}$, parameter $\lambda > 0$, 
            current tree $h$ with $I$ leaves, current leaf $i \in [I]$, 
            set of sorted thresholds $B_d$ for $d \in [D]$, 
            set of permutations $\sigma_d$ such that $x_{\sigma_d(1), d} \leq \dots \leq x_{\sigma_d(N), d}$ for $d \in [D]$, 
            and global variables $f_{n, i}, v_{n, i}, V_n$ for $n \in [N]$ and $i \in [I]$. 
        }
        \ENSURE{a best split condition $(d^\ast, b^\ast)$ and predictive labels $\hatyl^\ast, \hatyr^\ast$. }
        \STATE /* Initialize Terms for Computing Objective Value */
        \STATE $\bar{N} \leftarrow \sum_{n=1}^{N} (1 - f_{n, i}) \cdot \ind{h(\bx_n) \not= y_n}$; 
        \STATE $N^+ \leftarrow \sum_{n=1}^{N} f_{n, i} \cdot \ind{y_n = +1}$; $N^- \leftarrow \sum_{n=1}^{N} f_{n, i} \cdot \ind{y_n = -1}$; 
        \STATE $M \leftarrow \sum_{n=1}^{N} \omega_{n, i}$; $\bar{M} \leftarrow \sum_{n=1}^{N} \omega_{n, i} \cdot v_{n, i}$; 
        \STATE $\omega_{n, i} \leftarrow \ind{V_{n} - v_{n, i} \cdot \ind{\haty_i = +1} = 0} \; (\forall n \in [N])$; 
        \STATE /* Search All Features and Thresholds */
        \FOR{$d = 1, 2, \dots, D$}
            \STATE $N^+_\mathrm{L} \leftarrow 0$; $N^+_\mathrm{R} \leftarrow N_+$; $N^-_\mathrm{L} \leftarrow 0$; $N^-_\mathrm{R} \leftarrow N_-$; $M_\mathrm{L} \leftarrow 0$; $M_\mathrm{R} \leftarrow \bar{M}$; 
            \STATE $n, n_\mathrm{L}, n_\mathrm{R} \leftarrow \sigma_d(1)$; $m, m_\mathrm{L}, m_\mathrm{R} \leftarrow 1$; 
            \FOR{$b \in B_d$}
                \STATE /* Update Terms for Computing Empirical Risk */
                \WHILE{$x_{n, d} \leq b$ and $m \leq N$}
                    \STATE $N^+_\mathrm{L}, N^-_\mathrm{L} \leftarrow N^+_\mathrm{L} + f_{n, i} \cdot \ind{y_n = +1}, N^-_\mathrm{L} + f_{n, i} \cdot \ind{y_n = -1}$; 
                    \STATE $N^+_\mathrm{R}, N^-_\mathrm{R} \leftarrow N^+_\mathrm{R} - f_{n, i} \cdot \ind{y_n = +1}, N^-_\mathrm{R} - f_{n, i} \cdot \ind{y_n = -1}$; 
                    \STATE $m \leftarrow m + 1$; $n \leftarrow \sigma_d(m)$; 
                \ENDWHILE
                \STATE /* Update Terms for Computing Empirical Recourse Risk */
                \WHILE{$g_{n_\mathrm{L}}(d, b) = 1$ and $m_\mathrm{L} \leq N$}
                    \STATE $M_\mathrm{L} \leftarrow M_\mathrm{L} + \omega_{n_\mathrm{L}, i} \cdot v_{n_\mathrm{L}, i}$; $m_\mathrm{L} \leftarrow m_\mathrm{L} + 1$; $n_\mathrm{L} \leftarrow \sigma_d(m_\mathrm{L})$; 
                \ENDWHILE
                \WHILE{$\bar{g}_{n_\mathrm{R}}(d, b) = 0$ and $m_\mathrm{R} \leq N$}
                    \STATE $M_\mathrm{R} \leftarrow M_\mathrm{R} - \omega_{n_\mathrm{R}, i} \cdot v_{n_\mathrm{R}, i}$; $m_\mathrm{R} \leftarrow m_\mathrm{R} + 1$; $n_\mathrm{R} \leftarrow \sigma_d(m_\mathrm{R})$; 
                \ENDWHILE
                \STATE /* Optimize Predictive Labels using \cref{eq:err,eq:er} */
                \STATE $\hatyl(d, b), \hatyr(d, b) \leftarrow \argmin_{\hatyl, \hatyr \in \mathcal{Y}} \Phi_\lambda(d, b, \hatyl, \hatyr)$; \label{line:inner}
            \ENDFOR
        \ENDFOR
        \STATE /* Determine Best Split Condition */
        \STATE $(d^\ast, b^\ast) \leftarrow \argmax_{d \in [D], b \in B_d} \Phi_\lambda(d, b, \hatyl(d, b), \hatyr(d, b))$; \label{line:outer}
        \STATE /* Update Global Variables */
        \FOR{$n = 1, 2, \dots, N$}
            \STATE $f_{n, \mathrm{L}} \leftarrow f_{n, i} \cdot \ind{x_{n, d^\ast} \leq b^\ast}$; $f_{n, \mathrm{R}} \leftarrow f_{n, i} \cdot \ind{x_{n, d^\ast} > b^\ast}$; 
            \STATE $v_{n, \mathrm{L}} \leftarrow v_{n, i} \cdot g_n(d^\ast, b^\ast)$; $v_{n, \mathrm{R}} \leftarrow v_{n, i} \cdot \bar{g}_n(d^\ast, b^\ast)$; 
            \STATE $V_n \leftarrow V_n - v_{n, i} \ind{\haty_{i} = +1} + v_{n, \mathrm{L}} \ind{\hatyl^\ast = +1} + v_{n, \mathrm{R}} \ind{\hatyr^\ast = +1}$; 
        \ENDFOR
        \STATE \textbf{return} $d^\ast, b^\ast, \hatyl(d^\ast, b^\ast), \hatyr(d^\ast, b^\ast)$; 
    \end{algorithmic}
    \label{algo:ract}
\end{algorithm}

Let $v_n(r) = \ind{\exists \ba \in \calA_\epsilon(\bx_n): \bx_n + \ba \in r}$ be the indicator whether the instance $\bx_n$ can reach the subspace $r$ by some action $a \in \calA_\epsilon(\bx_n)$. 
We also denote the set of leaves that $\bx_n$ can reach their corresponding subspaces by $\mathcal{I}_n = \set{i' \in [I] \mid v_n(r_{i'}) = 1}$. 
Recall that our recourse loss is defined by $l_\mathrm{rec}(\bx; h) = {\min}_{\ba \in \calA_{\epsilon}(\bx)} l_{01}(+1, h(\bx + \ba))$. 
Using $\mathcal{I}_n$, we can express the recourse loss of $h$ for $\bx_n$ as $l_\mathrm{rec}(\bx_n; h) = \min_{i' \in \mathcal{I}_n} l_{01}(+1, \haty_{i'})$; that is, if there is no leaf $i' \in [I]$ such that $i' \in \mathcal{I}_n$ and $\haty_{i'} = +1$, our recourse loss takes $1$; otherwise, it takes $0$. 
Based on this observation, we can express the recourse loss after splitting at $i$ as
\begin{align*}
    l_\mathrm{rec}(\bx_n; h') = \min \set{\omega_{n, i}, &1 - v_n(r_\mathrm{L}) \cdot \ind{\hatyl = +1}, \\ &1 - v_n(r_\mathrm{R}) \cdot \ind{\hatyr = +1}}, 
\end{align*}
where $\omega_{n, i} = \ind{\forall i' \in \mathcal{I}_n \setminus \set{i}: \haty_{i'} \not= +1}$ and $r_\mathrm{L} = \set{\bx \in \mathcal{X} \mid \bx \in r_i \land x_d \leq b}$ (resp.\ $r_\mathrm{R} = \set{\bx \in \mathcal{X} \mid \bx \in r_i \land x_d > b}$) is the subspace of the left (resp.\ right) child node of $i$.

\begin{figure*}[t]
    \centering
    \includegraphics[width=\textwidth]{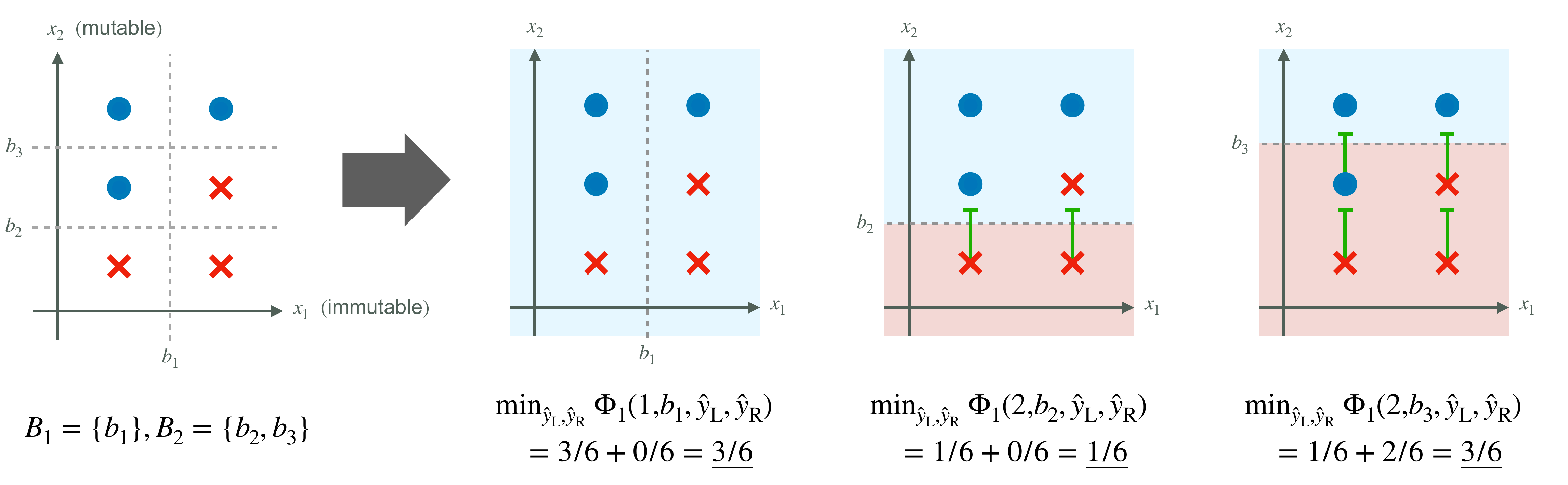}
    \vskip -0.1in
    \caption{
        Running example of our top-down greedy splitting algorithm on $\mathcal{X} = \mathbb{R}^2$.  
        Here, we have six labeled instances $(\bx, y)$ as a training sample $S$, and blue (resp.\ red) indicates the desired class $y = +1$ (reps. undesired class \ $y = -1$). 
        We assume that while the feature $x_1$ is immutable (i.e., can not be changed by an action), the feature $x_2$ is mutable (i.e., can be changed by an action). 
        We also assume thresholds $B_1 = \set{b_1}$ and $B_2 = \set{b_2, b_3}$ for the features $x_1$ and $x_2$, respectively, and set the trade-off parameter as $\lambda = 1$. 
        For each instance $\bx$ that is located in the red region corresponding to $\haty = -1$, the green line stands for the range where $\bx$ can reach by some action $\ba$ within a pre-defined cost budget $\epsilon$. 
        For each split condition $(d, b)$, we first determine predictive labels $\hatyl$ and $\hatyr$ by solving the inner problem of \eqref{eq:split}, and then find a best split condition that minimizes our objective function $\Phi_{\lambda}$. 
        Note that the objective value of each split condition $(d, b)$ can be computed in amortized constant time. 
    }
    \label{fig:alg:splitting}
\end{figure*}

To efficiently evaluate $\hat{\Omega}_\varepsilon(h' \mid S)$ for each $(d, b)$, we introduce some global variables and manage them, as with \cite{Guo:ICML2022}. 
Let $f_{n, i} = \mathbb{I}[\bx_n \in r_i]$ and $v_{n, i} = v_n(r_i)$. 
We denote the total number of leaves such that $v_{n, i'} = 1$ and $\haty_{i'} = +1$ by $V_n = \sum_{i' \in [I]} v_{n, i'} \cdot \ind{\haty_i = +1}$. 
By definition, $\omega_{n, i} = 1$ holds if and only if $V_n - v_{n, i} \cdot \ind{\haty_i = +1} = 0$. 
Using these global variables, we define four terms
\begin{align*}
    &M = {\sum}_{n=1}^{N} \omega_{n, i}, \;\; \bar{M} = {\sum}_{n=1}^{N} \omega_{n, i} \cdot v_{n, i}, \\
    &M_\mathrm{L} = {\sum}_{n=1}^{N} \omega_{n, i} \cdot v_{n, i} \cdot g_n(d, b), \\
    &M_\mathrm{R} = {\sum}_{n=1}^{N} \omega_{n, i} \cdot v_{n, i} \cdot \bar{g}_n(d, b),
\end{align*}
where $g_n(d, b) = \mathbb{I}[x_{n, d} + {\min}_{a \in \calA_\varepsilon(\bx_n)} a_d \leq b]$ and $\bar{g}_n(d, b) = \mathbb{I}[x_{n, d} + {\max}_{a \in \calA_\varepsilon(\bx_n)} a_d > b]$. 
From our assumption on the cost function $c$ and the definitions of $r_\mathrm{L}$ and $r_\mathrm{R}$, we have $v_n(r_\mathrm{L}) = v_{n, i} \cdot g_n(d, b)$ and $v_n(r_\mathrm{R}) = v_{n, i} \cdot \bar{g}_n(d, b)$~\cite{Guo:ICML2022}. 
Note that we can compute $g_n(d, b)$ and $\bar{g}_n(d, b)$ as constants in advance when $\bx_{n}, \calA_{\epsilon}(\bx_n)$, and $B_d$ are given. 
Using $M, \bar{M}, M_\mathrm{L}, M_\mathrm{R}$, we can write the empirical recourse risk of $h'$ as
\begin{align}\label{eq:err}
    \hat{\Omega}_\epsilon(h' \mid S) =
    \begin{cases}
        \frac{M}{N} & (\hatyl = -1 \land \hatyr = -1), \\
        \frac{M - M_\mathrm{L}}{N} & (\hatyl = +1 \land \hatyr = -1), \\
        \frac{M - M_\mathrm{R}}{N} & (\hatyl = -1 \land \hatyr = +1), \\
        \frac{M - \bar{M}}{N} & (\hatyl = +1 \land \hatyr = +1), 
    \end{cases}
\end{align}
which can be verified by the definition of $\hat{\Omega}_\epsilon$ and $h'$. 
By definition, $M_\mathrm{L}$ and $M_\mathrm{R}$ depend on the split condition $(d, b)$. 
While a naive computation of $M_\mathrm{L}$ and $M_\mathrm{R}$ requires $\mathcal{O}(N)$ time for each $(d, b)$, we can compute them in amortized constant time using the monotonicity of $g_n(d, b)$ and $\bar{g}_n(d, b)$ with respect to $n$ and $b$, which is described in \cref{appendix:proof:algo}.

\cref{algo:ract} presents an algorithm for the problem \eqref{eq:split}. 
It first determines predictive labels $\hatyl$ and $\hatyr$ by solving its inner problem for each $(d, b)$. 
As with the empirical recourse risk, we can also express the empirical risk of $h'$ by 
\begin{align}\label{eq:er}
    \hat{R}(h' \mid S) =
    \begin{cases}
        \frac{N^+ + \bar{N}}{N} & (\hatyl = -1 \land \hatyr = -1), \\
        \frac{N^-_\mathrm{L} + N^+_\mathrm{R} + \bar{N}}{N} & (\hatyl = +1 \land \hatyr = -1), \\
        \frac{N^+_\mathrm{L} + N^-_\mathrm{R} + \bar{N}}{N} & (\hatyl = -1 \land \hatyr = +1), \\
        \frac{N^- + \bar{N}}{N} & (\hatyl = +1 \land \hatyr = +1), 
    \end{cases}
\end{align}
and compute the terms for each $(d, b)$ in amortized constant time. 
After optimizing $\hatyl$ and $\hatyr$ for all $(d, b)$, \cref{algo:ract} determines a best split condition $(d^\ast, b^\ast)$ that minimizes $\Phi_\lambda$. 
In \cref{prop:algo}, we show that our algorithm for solving the problem~\eqref{eq:split} runs in the same time complexity as the standard learning algorithm for classification trees. 

\begin{proposition}\label{prop:algo}
    \cref{algo:ract} solves the problem \eqref{eq:split} in $\mathcal{O}(D \cdot N)$. 
\end{proposition}

\cref{fig:alg:splitting} presents a running example of \cref{algo:ract}. 
After determining a best split condition $(d^\ast, b^\ast)$ and predictive labels $\hatyl^\ast, \hatyr^\ast$ by \cref{algo:ract}, we apply \cref{algo:ract} to the left and right children with the updated global variables. 
We grow a tree by recursively repeating this procedure until some termination condition is met (e.g., maximum depth). 
Note that we need to compute the permutations $\sigma_d$ and indicators $g_n(d, b)$ and $\bar{g}_n(d, b)$ for $n \in [N], d \in [D], b \in B_d$, which roughly take $\mathcal{O}(D \cdot N \cdot \log N)$ and $\mathcal{O}(D \cdot N^2)$, respectively. 
However, we can compute them once as preprocessing, and the computation time of growing a tree was often more dominant than that of such preprocessing in practice.

\subsection{Set-Cover Relabeling}
While minimizing the objective function $\Phi_\lambda$ of the problem \eqref{eq:split} encourages the existence of recourse actions, the learned classification tree $\hat{h}$ does not necessarily satisfy the constraint $\hat{\Omega}_{\epsilon}(\hat{h} \mid S) \leq \delta$ of \cref{prob:tears}. 
We now introduce a post-processing task, called \emph{relabeling}~\cite{Kamiran:ICDM2010}, that modifies the predictive labels of leaves in $\hat{h}$ so as to satisfy the constraint. 
We also show that the task can be reduced to a variant of the \emph{minimum set cover} problem, which is known to be efficiently solved by a greedy algorithm with an approximation guarantee~\cite{Chvatal:MOR1979,Kearns:1990:computational}. 

Let $\hat{I}$ be the total number of leaves in a learned classification tree $\hat{h}$. 
To satisfy the constraint, a trivial solution is to set all the predictive labels of $\hat{h}$ to $+1$. 
However, it means to make $\hat{h}$ a constant classifier, i.e., $\hat{h}(\bx) = +1$ for any $\bx \in \mathcal{X}$, and may significantly degrade the predictive performance. 
Thus, our aim is to determine the predictive label $\haty_i$ of each leaf $i \in [\hat{I}]$ so as to satisfy the constraint $\hat{\Omega}_{\epsilon}(\hat{h} \mid S) \leq \delta$ without increasing the empirical risk $\hat{R}(\hat{h} \mid S)$ as much as possible. 

We can minimize $\hat{R}$ by setting $\haty_i$ as the majority class among the instances $\bx_n$ such that $\bx_n \in r_i$. 
In addition, our key observation on $\hat{\Omega}_\epsilon$ is that we can express it as $\hat{\Omega}_{\epsilon}(h \mid S) = 1 - \frac{1}{N} |\bigcup_{i \in [I]: \haty_i = +1} \mathcal{N}_i|$, where $\mathcal{N}_i \coloneqq \set{n \in [N] \mid v_n(r_i)}$ is the set of instances $\bx_n$ that can reach the leaf $i$ by some action $\ba \in \calA_\epsilon(\bx_n)$. 
It implies that we can reduce $\hat{\Omega}_{\epsilon}(h \mid S)$ only by changing the predictive labels of the leaves $i$ with $\haty_i = -1$ to $+1$. 
Based on these observations, we first initialize each predictive label as $\haty_i = 1 - 2 \cdot \ind{N^+_i < N^-_i}$, where $N^+_i = |\set{n \in [N] \mid \bx_n \in r_i \land y_n = +1}|$ and $N^-_i = |\set{n \in [N] \mid \bx_n \in r_i \land y_n = -1}|$. 
Then, we select leaves $\mathcal{I} \subseteq [\hat{I}]$ that should be changed from $\haty_i = -1$ to $\haty_i = +1$ so as to minimize the increase of $\hat{R}(\hat{h} \mid S)$ under the constraint $\hat{\Omega}_{\epsilon}(\hat{h} \mid S) \leq \delta$. 
By ignoring constant terms, this task can be formulated as 
\begin{equation}
    \label{eq:cover}
        \min_{\mathcal{I} \subseteq \mathcal{I}^-} \; \frac{1}{N} {\sum}_{i \in \mathcal{I}} c_i \;\;\;
        \text{s.t.} \;\; \frac{1}{N} \left|{\bigcup}_{i \in \mathcal{I}^+ \cup \mathcal{I}} \mathcal{N}_i \right| \geq 1 - \delta, 
\end{equation}
where $\mathcal{I}^- = \set{i \in [\hat{I}] \mid \haty_i = -1}$, $\mathcal{I}^+ = [\hat{I}] \setminus \mathcal{I}^-$, and $c_i = N^-_i - N^+_i$. 
By modifying the predictive labels of the leaves in a feasible solution $\mathcal{I}$ to the problem \eqref{eq:cover} as $+1$, we can obtain a classification tree $h^\ast$ that satisfies $\hat{\Omega}_{\epsilon}(h^\ast \mid S) \leq \delta$. 
In \cref{prop:wpc}, we show that the problem \eqref{eq:cover} belongs to a class of a well-known combinatorial optimization problem that can be efficiently solved with an approximation guarantee~\cite{Chvatal:MOR1979}. 

\begin{proposition}\label{prop:wpc}
    The problem~\eqref{eq:cover} is reduced to the weighted partial cover problem. 
\end{proposition}

While the weighted partial cover problem is NP-hard, there exist polynomial-time approximation algorithms (e.g., a greedy algorithm achieves $(2 \cdot H(N) + 3)$-approximation, where $H(N) = \sum_{n=1}^{N} 1/n = \Theta(\ln N)$~\cite{Kearns:1990:computational}).

\paragraph{PAC-Style Guarantee.}
By solving the problem  \eqref{eq:cover}, we can obtain a classifier $h^\ast$ that can provide valid actions to at least $100 \cdot (1 - \delta)$ \% instances in a given training sample $S$. 
However, since our empirical recourse risk $\hat{\Omega}_\varepsilon$ is the probability of the existence of valid actions estimated over $S$, there is no guarantee for unseen test instances.  
To analyze this risk, we show a PAC-style bound on the estimation error of $\hat{\Omega}_\varepsilon$ in \cref{prop:pac}~\cite{Mohri:2012:Foundations}. 

\begin{proposition}\label{prop:pac}
    For a classifier $h \in \mathcal{H}$, let $\Omega_\epsilon(h) \coloneqq \mathbb{P}_{x}[\exists \ba \in \calA_{\epsilon}(\bx): h(\bx + \ba) = +1]$ be the expected recourse risk of $h$. 
    Given a sample $S$ and $\alpha > 0$, the following inequality holds with probability at least $1 - \alpha$:
    \begin{align*}
        \Omega_\epsilon(h) \leq \hat{\Omega}_\epsilon(h \mid S) + \sqrt{\frac{\ln |\mathcal{H}| - \ln \alpha}{2 \cdot |S|}}.
    \end{align*}
\end{proposition}

\cref{prop:pac} implies that the estimation error of $\hat{\Omega}_\varepsilon$ depends mainly on the sizes of $\mathcal{H}$ and $S$. 
Since the problem~\eqref{eq:cover} considers only classification trees that are same with $\hat{h}$ except for the predictive labels of the leaves in $\mathcal{I}^-$, we have $|\mathcal{H}| = 2^{|\mathcal{I}^-|}$. 
Thus, by replacing $\delta$ in the problem~\eqref{eq:cover} with $\delta' = \delta - \sqrt{(|\mathcal{I}^-| \cdot \ln 2 - \ln \alpha)/(2 \cdot N)}$, we can obtain a classification tree $h^\ast$ that satisfies $\Omega_\epsilon(h^\ast) \leq \delta$ with probability at least $1 - \alpha$. 
Note that this condition is equivalent to the \emph{probably approximately has recourse (PARE)} condition proposed by~\cite{Ross:NIPS2021}.

\subsection{Extension to Random Forest}
To learn a tree ensemble classifier, we employ our RACT algorithm as a base learner of the random forest~\cite{Breiman:ML2001}. 
A random forest classifier $h$ makes predictions by majority voting of $T$ classification trees $h_1, \dots, h_T$, i.e., $h(\bx) = \operatorname{sgn}(\sum_{t=1}^{T} h_t(\bx))$. 
To apply RACT to the random forest framework, we can simply replace the algorithm for finding a best split condition of each $h_t$ with \cref{algo:ract}. 

\begin{figure*}[t]
    \centering
    \subfigure[Classification Tree]{
        \includegraphics[width=\linewidth]{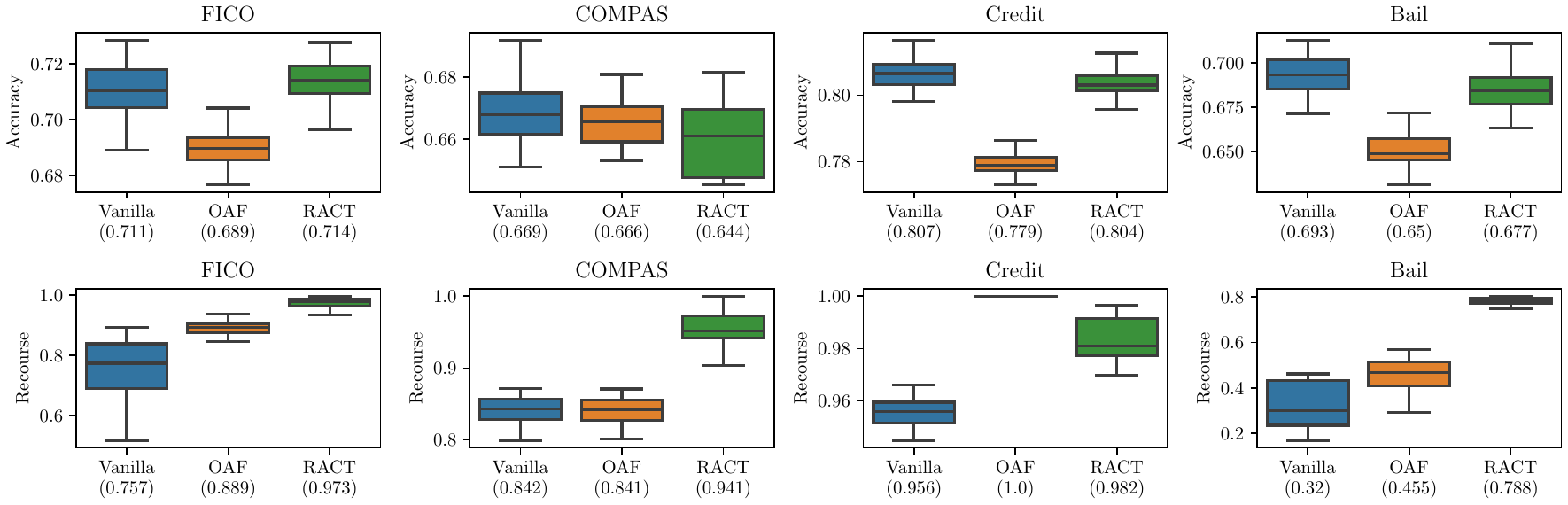}
        \label{fig:exp:comp:ct}
    }
    \subfigure[Random Forest]{
        \includegraphics[width=\linewidth]{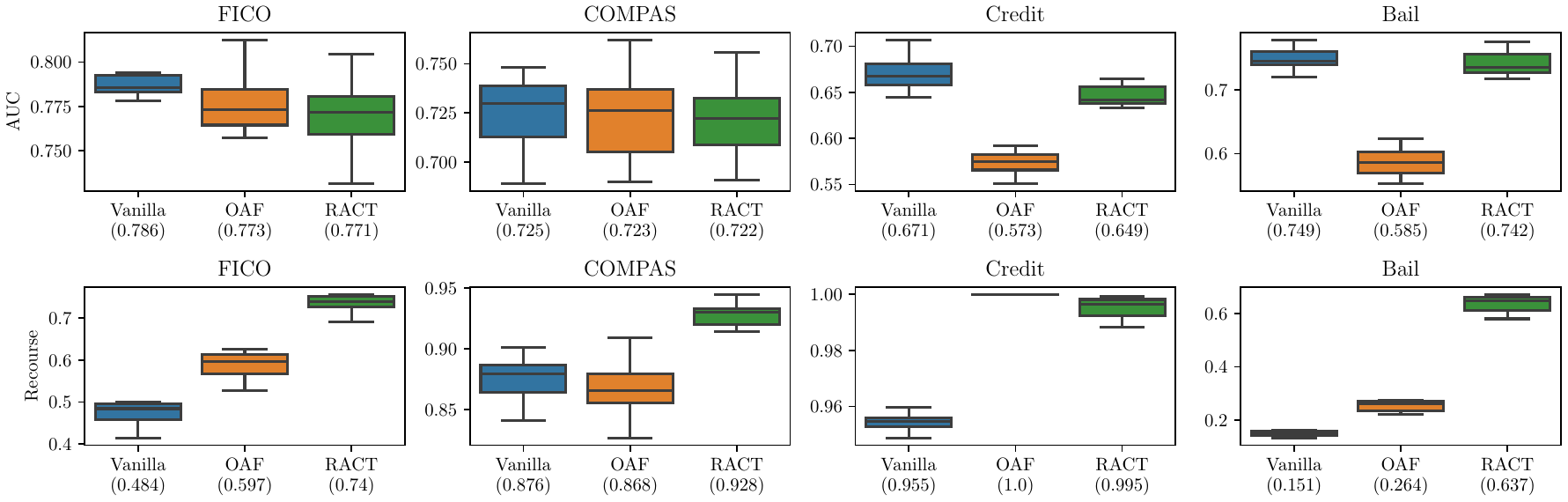}
        \label{fig:exp:comp:rf}
    }
    \caption{
        Experimental results of baseline comparison. 
        Figures in the top (resp.\ bottom) row present predictive accuracy (resp.\ recourse ratio). 
        Our RACT attained higher recourse ratio than the baselines while keeping comparable accuracy on almost all the datasets. 
    }
    \label{fig:exp:comp}
\end{figure*}

\section{Experiments}\label{sec:exp}
To investigate the performance of our RACT, we conducted experiments on real datasets. 
All the code was implemented in Python 3.10 with Numba 0.56.4 and is available at \url{https://github.com/kelicht/ract}. 
All the experiments were conducted on macOS Monterey with Apple M1 Pro CPU and 32 GB memory. 

Our experimental evaluation aims to answer the following questions:
(i)~How are the predictive accuracy and recourse guarantee of tree-based models learned by our RACT compared to those by the baselines? 
(ii)~How is the quality of the recourse actions extracted from the tree-based models in the sense of their practicality? 
(iii)~Can our RACT balance the trade-off between accuracy and recourse? 
Due to page limitations, the complete results are shown in \cref{appendix:exp}. 

\paragraph{Experimental Settings. }
We used four real datasets: FICO~($N=9871, D=23$)~\cite{fico:2018}, COMPAS~($N=6167, D=14$)~\cite{compas:2016}, Credit~($N=30000, D=16$)~\cite{Credit:2009}, and Bail~($N=8923, D=16$)~\cite{Bail:1988}. 
As a cost function, we used the \emph{max percentile shift (MPS)}~\cite{Ustun:FAT*2019} defined as 
\begin{align*}
    c(\ba \mid \bx) = \max_{d \in [D]} \left| Q_d(x_d + a_d) - Q_d(x_d) \right|,    
\end{align*}
where $Q_d$ is the cumulative distribution function of a feature $d$. 
To optimize an action $\ba$ for each instance $\bx$ by solving the problem \eqref{eq:ce}, we employed the \emph{feature tweaking algorithm}~\cite{Tolomei:KDD2017}, which is a fast heuristic algorithm for tree-based models. 
Note that we also employed the exact method based on integer optimization~\cite{Cui:KDD2015,Kanamori:IJCAI2020}, and its results are presented in \cref{appendix:exp}. 
To the best of our knowledge, there is no existing method for learning tree-based models while guaranteeing the existence of actions. 
Thus, we compared our RACT with two baselines: learning with no constraint on recourse (\emph{Vanilla}) and learning with only actionable features (\emph{OAF}), following \cite{Dominguez-olmedo:ICML2022}.

\subsection{Performance Comparison}\label{sec:exp:comp}
First, we evaluate the performance of classification trees and random forest classifiers learned by our RACT in comparison with the baselines. 
We conducted $10$-fold cross validation, and measured (i)~the average accuracy and AUC on the test set, (ii)~the average recourse ratio, which is defined as the ratio of the test instances that are guaranteed valid actions whose costs are less than $\epsilon = 0.3$, and (iii)~the average running time. 
For the baselines and our RACT, we trained classification trees with a maximum depth of $64$ and random forest classifiers with $T=200$ classification trees. 
For each dataset, we determined the hyper-parameters $\delta$ and $\lambda$ of our RACT based on the results of our trade-off analyses in \cref{sec:exp:tradeoff}. 
Specifically, we first drew scatter plots between the average accuracy and recourse ratio of the model trained with each $\delta$ and $\lambda$, and selected one of the Pareto-optimal solutions that achieves a good balance between the accuracy and recourse ratio as $\delta$ and $\lambda$. 

\begin{table}[t]
    \centering
    \scriptsize
    \caption{
        Experimental results on the average running time [s] for random forest classifiers. 
        There is no significant difference between Vanilla and our RACT on almost all the datasets. 
    }
\vskip 0.1in
    \begin{tabular}{cccc}
    \toprule
        \textbf{Dataset} & \textbf{Vanilla} & \textbf{OAF} & \textbf{RACT} \\
    \midrule
        FICO & $42.09 \pm 1.8$ & $35.25 \pm 1.52$ & $48.07 \pm 2.33$ \\
        COMPAS & $2.31 \pm 0.32$ & $1.76 \pm 0.14$ & $2.07 \pm 0.13$ \\
        Credit & $292.53 \pm 21.6$ & $268.41 \pm 18.9$ & $277.40 \pm 25.8$ \\
        Bail & $24.84 \pm 1.93$ & $3.26 \pm 0.35$ & $23.04 \pm 1.64$ \\
    \bottomrule
    \end{tabular}
    \label{tab:exp:comp:time:rf}
\end{table}

\cref{fig:exp:comp} presents the results on the predictive accuracy and recourse ratio. 
From \cref{fig:exp:comp}, we observed that
(i)~our RACT attained comparable predictive accuracy to the baselines on almost all the datasets, and 
(ii)~our RACT achieved significantly higher recourse ratios than the baselines except for OAF on the Credit dataset\footnote{We would like to note that OAF on the Credit dataset obtained a constant classifier that predicts any input instance as the desired class for all the folds, which is the reason why OAF achieved $100\%$ recourse ratio on the Credit dataset in \cref{fig:exp:comp}. }. 
These results suggest that our RACT achieved higher recourse ratios than the baselines while keeping comparable accuracy. 
\cref{tab:exp:comp:time:rf} shows the average running time for random forests. 
We can see that there is no significant difference in the running time between Vanilla and RACT, which indicates that our algorithm that considers the additional constraint on the empirical recourse risk performed as fast as the existing one without the constraint. 
In summary, we have confirmed that \emph{our method succeeded in guaranteeing the existence of recourse actions for more instances than the baselines without compromising predictive accuracy and computational efficiency}.

\begin{table}[t]
    \centering
    \scriptsize
    \caption{
        Average cost of extracted actions (lower is better). 
        We used the MPS~\cite{Ustun:FAT*2019} as a cost function $c$. 
        Our RACT attained lower costs than the baselines regardless of the datasets. 
    }
\vskip 0.1in
    \begin{tabular}{cccc}
    \toprule
        \textbf{Dataset} & \textbf{Vanilla} & \textbf{OAF} & \textbf{RACT} \\
    \midrule
        FICO & $0.447 \pm 0.05$ & $0.407 \pm 0.03$ & $\bm{0.283 \pm 0.01}$ \\
        COMPAS & $0.298 \pm 0.02$ & $0.28 \pm 0.01$ & $\bm{0.232 \pm 0.02}$ \\
        Credit & $0.293 \pm 0.02$ & N/A & $\bm{0.166 \pm 0.04}$ \\
        Bail & $0.763 \pm 0.03$ & $0.525 \pm 0.04$ & $\bm{0.419 \pm 0.05}$ \\
    \bottomrule
    \end{tabular}
    \label{tab:exp:qual:cost}
\end{table}
\begin{table}[t]
    \centering
    \scriptsize
    \caption{
        Average plausibility of extracted actions (lower is better). 
        Following \cite{Parmentier:ICML2021}, we measured the outlier score estimated by isolation forests. 
        There is no significant difference in the plausibility between the baselines and our RACT. 
    }
\vskip 0.1in
    \begin{tabular}{cccc}
    \toprule
        \textbf{Dataset} & \textbf{Vanilla} & \textbf{OAF} & \textbf{RACT} \\
    \midrule
        FICO & $0.456 \pm 0.0$ & $0.446 \pm 0.0$ & $\bm{0.437 \pm 0.0}$ \\
        COMPAS & $\bm{0.44 \pm 0.01}$ & $0.447 \pm 0.01$ & $0.453 \pm 0.01$ \\
        Credit & $0.526 \pm 0.01$ & N/A & $\bm{0.523 \pm 0.01}$ \\
        Bail & $\bm{0.504 \pm 0.01}$ & $0.507 \pm 0.0$ & $0.512 \pm 0.01$ \\
    \bottomrule
    \end{tabular}
    \label{tab:exp:qual:plausibility}
\end{table}
\begin{table}[t]
    \centering
    \scriptsize
    \caption{
        Average validity of extracted actions under the causal recourse constraint proposed by \citet{Karimi:FAccT2021} (higher is better). 
        Our RACT attained higher validity than the baselines even in the causal recourse setting. 
    }
\vskip 0.1in
    \begin{tabular}{cccc}
    \toprule
        \textbf{Dataset} & \textbf{Vanilla} & \textbf{OAF} & \textbf{RACT} \\
    \midrule
        FICO & $0.316 \pm 0.04$ & $0.502 \pm 0.03$ & $\bm{0.648 \pm 0.03}$ \\
        COMPAS & $0.599 \pm 0.02$ & $0.642 \pm 0.04$ & $\bm{0.721 \pm 0.02}$ \\
        Credit & $0.669 \pm 0.05$ & N/A & $\bm{0.964 \pm 0.03}$ \\
        Bail & $0.174 \pm 0.04$ & $0.201 \pm 0.04$ & $\bm{0.539 \pm 0.03}$ \\
    \bottomrule
    \end{tabular}
    \label{tab:exp:qual:causality}
\end{table}

\subsection{Recourse Quality Analysis}\label{sec:exp:qual}
Next, we examine the quality of extracted actions from tree-based models learned by our RACT. 
To assess the practicality of the actions, we evaluate
(i)~their average \emph{cost}~\cite{Ustun:FAT*2019}, 
(ii)~their \emph{plausibility}~\cite{Parmentier:ICML2021}, and 
(iii)~their ratio that they are valid, which we call validity, in the \emph{causal recourse} setting~\cite{Karimi:FAccT2021}. 
We report the results for random forest classifiers here. 

\paragraph{Cost. }
\cref{tab:exp:qual:cost} shows the average cost $c(\ba \mid \bx)$ of the obtained valid actions $\ba$ for test instances $\bx$. 
From \cref{tab:exp:qual:cost}, we see that RACT attained the lowest costs regardless of the datasets, which suggests that our RACT succeeded in providing affected individuals with more efficient and executable actions than those of the baselines. 

\paragraph{Plausibility. }
\cref{tab:exp:qual:plausibility} shows the average plausibility of the obtained valid actions $\ba$ for test instances $\bx$. 
To evaluate the plausibility of $\ba$ for $\bx$, the previous studies often use the outlier score of $\bx + \ba$~\cite{Parmentier:ICML2021}. 
Following the previous study, we employed isolation forests~\cite{Liu:ICDM2008} for estimating the outlier score. 
From \cref{tab:exp:qual:plausibility}, we observed that there is no significant difference in plausibility between the baselines and RACT. 
These results indicate that our RACT did not harm the plausibility of actions, and its provided actions are as reasonable as those of the baselines. 

\paragraph{Causal Recourse. }
\cref{tab:exp:qual:causality} presents the average validity of the obtained actions for test instances in the causal recourse setting~\cite{Karimi:FAccT2021}. 
For each dataset, we estimated causal DAGs by DirectLiNGAM~\cite{Shimizu:JMLR2011}, which is a popular method for estimating causal DAGs, and optimized actions under the causal recourse constraint with the estimated causal DAGs. 
From \cref{tab:exp:qual:causality}, we see that our RACT attained higher validity than the baselines. 
These results indicate that our RACT succeeded in guaranteeing the existence of valid actions for more individuals than the baselines even in the causal recourse setting.  

To conclude, we have confirmed that \emph{our method could provide affected individuals with practical recourse actions in the sense of their cost, plausibility, and causality}.

\begin{figure*}[t]
    \centering
    \includegraphics[width=\linewidth]{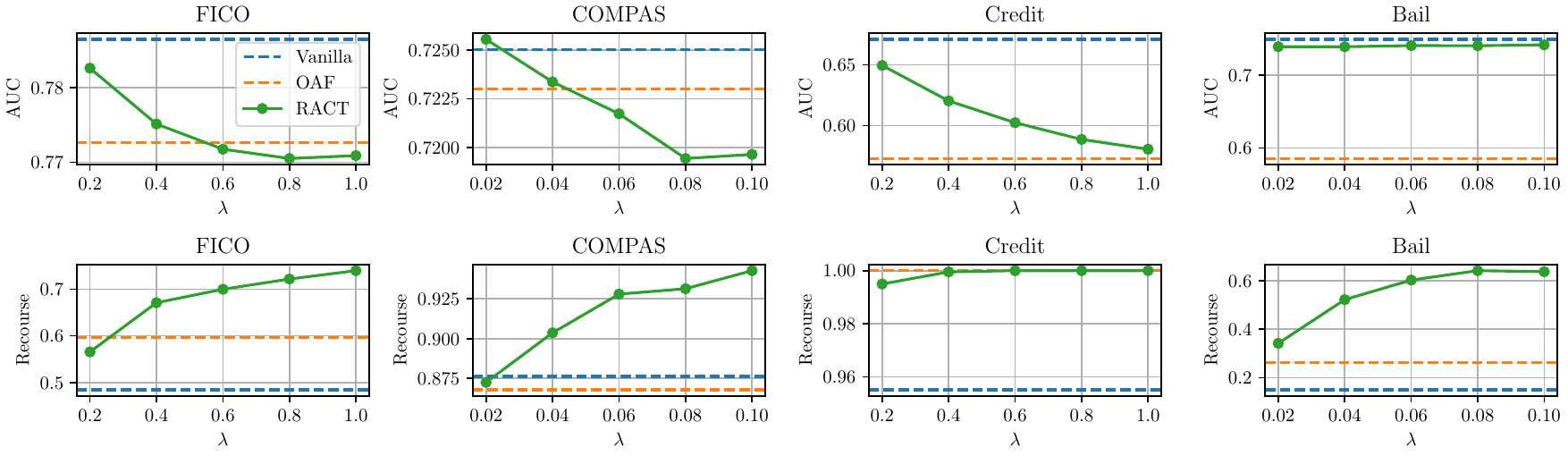}
    \vskip -0.1in
    \caption{
        Sensitivity analyses of the trade-off parameter $\lambda$ with respect to the average AUC and recourse ratio. 
        While the value ranges of $\lambda$ are different across the datasets, we can see that the recourse ratio (resp.\ AUC) was improved as $\lambda$ increased (resp.\ decreased) overall. 
    }
    \label{fig:exp:tradeoff:forest}
\end{figure*}
\subsection{Trade-Off Analysis}\label{sec:exp:tradeoff}
Finally, we analyze the trade-off between the predictive accuracy and recourse guarantee of our RACT by varying its trade-off parameters. 
As with \cref{sec:exp:qual}, we report the results for random forest classifiers here. 
We trained random forest classifiers by varying the trade-off parameter $\lambda$, and measured their average AUC and recourse ratio. 

\cref{fig:exp:tradeoff:forest} shows the results for each $\lambda$. 
From \cref{fig:exp:tradeoff:forest}, we see that the recourse ratio (resp.\ AUC) was improved by increasing (resp.\ decreasing) $\lambda$ on almost all the datasets. 
These results suggest not only that we can attain a desired trade-off by tuning $\lambda$, but also that we have a chance to achieve a higher recourse ratio than the baselines while keeping comparable accuracy by choosing an appropriate value of $\lambda$. 
In summary, we have confirmed that \emph{our method can balance the trade-off between the predictive accuracy and recourse guarantee by tuning the parameter $\lambda$}.

\section{Conclusion}
In this paper, we proposed a new framework for learning tree-based models, named recourse-aware classification tree (RACT), that can provide both accurate predictions and executable recourse actions. 
We first introduced the recourse risk to evaluate the ratio of instances that are not guaranteed to have recourse actions, and formulated a task of learning a classification tree with the constraint on the recourse risk. 
Then, we proposed an efficient top-down greedy learning algorithm by leveraging adversarial training techniques, and showed that we can modify a learned tree so as to satisfy the constraint by solving a well-known variant of the minimum set-cover problem. 
We also showed that our algorithm can be easily applied to the random forest, which is one of the popular frameworks for learning tree ensembles. 
Experimental results demonstrated that our RACT succeeded in guaranteeing recourse actions for more individuals than the baselines while keeping comparable accuracy and efficiency. 

\paragraph{Limitations and Future Work. }
There are several directions to improve our RACT. 
First, it is important to extend our RACT to the popular frameworks of gradient boosted trees, such as XGBoost~\cite{Chen:KDD2016} and LightGBM~\cite{Ke:NIPS2017}. 
For that purpose, we need to extend our algorithm to learning regression trees~\cite{Andriushchenko:NIPS2019}. 
However, it is not trivial to develop such methods without degrading computational efficiency. 
In addition, the computational efficiency of our proposed algorithm relies on our assumption of the $\ell_\infty$-type cost function $c$. 
While we can easily decide whether the budget constraint is violated or not by checking each feature independently for the $\ell_\infty$-type cost functions, such a property does not hold for general cost functions, including $\ell_1$- or $\ell_2$-type cost functions. 
We expect that when we introduce some heuristic strategies, such as changing the budget parameter $\varepsilon$ for each depth of tree~\cite{Wang:ICML2020}, our algorithm can be extended to deal with the $\ell_1$- and $\ell_2$-type cost functions with keeping efficiency. 
Finally, in our experiments of \cref{sec:exp:tradeoff}, we individually set the range of our trade-off parameter $\lambda$ to be searched for each dataset. 
Because such a procedure is costly in practice, we need to further analyze its sensitivity to determine its default range automatically.

\section*{Acknowledgments}
We thank the anonymous reviewers for their insightful comments. 
This work was supported in part by JST ACT-X JPMJAX23C6. 

\section*{Impact Statement}
\paragraph{Positive Impacts.}
Our proposed method, named RACT, is a new framework that aims to learn accurate tree-based models while guaranteeing the existence of recourse actions. 
As demonstrated in our experiments in \cref{sec:exp}, tree-based models trained by our method can provide executable recourse actions to more individuals than the existing methods without degrading predictive accuracy. 
Thus, our method enables us to learn tree-based models that make accurate predictions and guarantee recourse actions, which improves the trustworthiness of algorithmic decision-making for critical tasks in the real world such as loan approvals and judicial decisions~\cite{Karimi:ACMCS2022,Ross:NIPS2021}. 
In addition, our method gives decision-makers an insight into the trade-off between the predictive accuracy and recourse ratio by tuning the hyper-parameter $\lambda$~\cite{Levanon:ICML2021,Olckers:arxiv2023}. 

\paragraph{Negative Impacts.}
While our method has several positive impacts, there may exist some negative impacts as well. 
First, in practice, it is sometimes undesirable for decision-makers to guarantee the existence of executable actions for all individuals. 
For example, providing easy actions for granting loans to applicants who do not have the capacity to repay might cause a serious financial crisis in the future. 
However, since our method can adjust the ratio of individuals who are guaranteed to have executable recourse actions by tuning its hyper-parameter $\lambda$, it helps decision-makers provide recourse actions to appropriate individuals while keeping accurate decisions~\cite{Levanon:ICML2021,Olckers:arxiv2023}. 
Second, there is a risk that our method may be used maliciously to train a model that provides specific actions for occuring some undesired situations, such as discrimination. 
To avoid this risk, we may need to check the actions provided to affected individuals before deploying the models (e.g., using the existing frameworks for obtaining the global summary of recourse actions~\cite{Rawal:NIPS2020,Kanamori:AISTATS2022}).

\bibliography{ref}
\bibliographystyle{icml2024}

\newpage
\appendix
\onecolumn

\section{Omitted Proofs}
\subsection{Proof of \cref{prop:algo}}\label{appendix:proof:algo}
\begin{proof}[Proof of \cref{prop:algo}]
    To prove \cref{prop:algo}, we first prove the correctness of \cref{algo:ract}. 
    Then, we show the complexity of \cref{algo:ract}. 

    \paragraph{Correctness.}
    At first, we show that \cref{algo:ract} computes the empirical recourse risk $\hat{\Omega}_\epsilon(h' \mid S)$ for each split condition $(d, b)$ and predictive labels $\hatyl, \hatyr$ correctly. 
    Recall that we define $\omega_{n, i} = \ind{\forall i' \in \mathcal{I}_n \setminus \set{i}: \haty_{i'} \not= +1}$ and that we have $v_n(r_\mathrm{L}) = v_{n, i} \cdot g_n(d, b)$ and $v_n(r_\mathrm{R}) = v_{n, i} \cdot \bar{g}_n(d, b)$. 
    From the definitions of our recourse loss $l_\mathrm{rec}$ and decision stump $h'$, we have
    \begin{align*}
        \hat{\Omega}_\epsilon(h' \mid S) 
        &= \frac{1}{N} \sum_{n=1}^{N} \min \set{\omega_{n, i}, 1 - v_n(r_\mathrm{L}) \cdot \ind{\hatyl = +1}, 1 - v_n(r_\mathrm{R}) \cdot \ind{\hatyr = +1}} \\
        &= 
        \begin{cases}
            \frac{1}{N} \sum_{n=1}^{N} \omega_{n, i}, & (\hatyl = -1 \land \hatyr = -1) \\
            \frac{1}{N} \sum_{n=1}^{N} \omega_{n, i} \cdot (1 - v_n(r_\mathrm{L})) & (\hatyl = +1 \land \hatyr = -1) \\
            \frac{1}{N} \sum_{n=1}^{N} \omega_{n, i} \cdot (1 - v_n(r_\mathrm{R})) & (\hatyl = -1 \land \hatyr = +1) \\
            \frac{1}{N} \sum_{n=1}^{N} \omega_{n, i} \cdot (1 - v_n(r_i)) & (\hatyl = +1 \land \hatyr = +1)
        \end{cases} \\
        &= 
        \begin{cases}
            \frac{1}{N} \sum_{n=1}^{N} \omega_{n, i}, & (\hatyl = -1 \land \hatyr = -1) \\
            \frac{1}{N} \sum_{n=1}^{N} \omega_{n, i} \cdot (1 - v_{n, i} \cdot g_n(d, b)) & (\hatyl = +1 \land \hatyr = -1) \\
            \frac{1}{N} \sum_{n=1}^{N} \omega_{n, i} \cdot (1 - v_{n, i} \cdot \bar{g}_n(d, b)) & (\hatyl = -1 \land \hatyr = +1) \\
            \frac{1}{N} \sum_{n=1}^{N} \omega_{n, i} \cdot (1 - v_{n, i}) & (\hatyl = +1 \land \hatyr = +1) 
        \end{cases} \\
        &=
        \begin{cases}
            \frac{M}{N} & (\hatyl = -1 \land \hatyr = -1) \\
            \frac{M - M_\mathrm{L}}{N} & (\hatyl = +1 \land \hatyr = -1) \\
            \frac{M - M_\mathrm{R}}{N} & (\hatyl = -1 \land \hatyr = +1) \\
            \frac{M - \bar{M}}{N} & (\hatyl = +1 \land \hatyr = +1), 
        \end{cases}
    \end{align*}
    which shows the correctness of \cref{eq:err}. 
    Thus, we can compute $\hat{\Omega}_\epsilon(h' \mid S)$ correctly if we have computed the terms $M, \bar{M}, M_\mathrm{L}, M_\mathrm{R}$ for each $(d, b)$. 
    Since $M$ and $\bar{M}$ is independent to $(d, b)$, we can compute them before searching $(d, b)$ (line~4 of \cref{algo:ract}). 
    In contrast, since $M_\mathrm{L}$ and $M_\mathrm{R}$ depends on $(d, b)$, we need to compute them for each $(d, b)$. 

    Here, we show that we can compute $M_\mathrm{L}$ and $M_\mathrm{R}$ for a $j$-th split condition $(d, b_j)$ using the results $M'_\mathrm{L}$ and $M'_\mathrm{R}$ for the previous split condition $(d, b_{j-1})$. 
    For notational simplicity, we assume $x_{1, d} < \dots < x_{N, d}$. 
    From the definition of $g_n$, we can see the two monotonic properties on $n \in [N]$ and $j \in [J_d]$: $g_{n}(d, b_j) \geq g_{n'}(d, b_j)$ for any $n' > n$ and $g_n(d, b_{j}) \geq g_n(d, b_{j'})$ for any $j' < j$. 
    While the former implies that $g_n(d, b_j) = 0 \implies g_{n'}(d, b_j) = 0$ holds for any $n' > n$, the latter implies that $g_n(d, b_j) = 1 \implies g_n(d, b_{j'}) = 1$ holds for any $j' > j$. 
    Let $m = \min_{n \in [N]: g_n(d, b_{j-1})=0} n$ and $m'= \max_{n \in [N]: g_n(d, b_j)=1} n$. 
    Using the above properties, we have $M_\mathrm{L} - M'_\mathrm{L} = \sum_{n = m}^{m'} \omega_{n, i} \cdot v_{n, i} \iff M_\mathrm{L} = M'_\mathrm{L} + \sum_{n = m}^{m'} \omega_{n, i} \cdot v_{n, i}$. 
    Furthermore, we can see the similar monotonic properties of $\bar{g}_n(d, b)$ as well, and we have $M_\mathrm{R} = M'_\mathrm{R} - \sum_{n = \bar{m}}^{\bar{m}'} \omega_{n, i} \cdot v_{n, i}$, where $\bar{m} = \min_{n \in [N]: \bar{g}_n(d, b_{j-1})=1} n$ and $\bar{m}'= \max_{n \in [N]: \bar{g}_n(d, b_j)=0} n$. 
    In lines~18--20 and lines~21--23, \cref{algo:ract} updates the terms $M_\mathrm{L}$ and $M_\mathrm{L}$ using these facts. 

    Similarly, we can show that \cref{algo:ract} computes the empirical risk $\hat{R}(h' \mid S)$ for each split condition $(d, b)$ and predictive labels $\hatyl, \hatyr$ correctly. 
    For \cref{eq:er}, we define as 
    \begin{align*}
        &\hat{N} = {\sum}_{n=1}^{N} (1-\ind{\bx_n \in r_i}) \cdot l(y_n, h(\bx_n)) ), N^+ = {\sum}_{n=1}^{N} f_{n, i} \cdot \ind{y_n = +1}, N^- = {\sum}_{n=1}^{N} f_{n, i} \cdot \ind{y_n = +1}, \\
        &N^+_\mathrm{L} = {\sum}_{n=1}^{N} f_{n, i} \cdot \ind{x_{n, d} \leq b} \cdot \ind{y_n = +1}, N^-_\mathrm{L} = {\sum}_{n=1}^{N} f_{n, i} \cdot \ind{x_{n, d} \leq b} \cdot \ind{y_n = -1}, \\
        &N^+_\mathrm{R} = {\sum}_{n=1}^{N} f_{n, i} \cdot \ind{x_{n, d} > b} \cdot \ind{y_n = +1}, N^-_\mathrm{R} = {\sum}_{n=1}^{N} f_{n, i} \cdot \ind{x_{n, d} > b} \cdot \ind{y_n = -1}, 
    \end{align*}
    where $f_{n, i} = \ind{\bx_n \in r_i}$. 
    The terms $N^+_\mathrm{L}, N^-_\mathrm{L}, N^+_\mathrm{R}, N^-_\mathrm{R}$ depend on $(d, b)$.
    As with the empirical recourse risk, \cref{algo:ract} computes them using the results of the previous split condition in lines~12--16, which leverages the monotonic properties of $\ind{x_{n, d} \leq b}$ and $\ind{x_{n, d} > b}$. 
    Using the above terms, we have
    \begin{align*}
        \hat{R}(h' \mid S)
        &= \frac{1}{N} \sum_{n=1}^{N} ( \ind{\bx_n \in r_i} \cdot l(y_n, h(\bx_n; d, b, \hatyl, \hatyr)) + (1-\ind{\bx_n \in r_i}) \cdot l(y_n, h(\bx_n)) ) \\
        &= \frac{1}{N} \sum_{n=1}^{N} ( f_{n, i} \cdot \ind{x_{n, d} \leq b} \cdot l(y_n, \hatyl) + f_{n, i} \cdot \ind{x_{n, d} > b} \cdot l(y_n, \hatyr) ) + \hat{N} \\
        &= 
        \begin{cases}
            \frac{1}{N} \sum_{n=1}^{N} f_{n, i} \cdot \ind{y_n = +1} + \hat{N}, & (\hatyl = -1 \land \hatyr = -1) \\
            \frac{1}{N} \sum_{n=1}^{N} (f_{n, i} \cdot \ind{x_{n, d} \leq b} \cdot \ind{y_n = -1} + f_{n, i} \cdot \ind{x_{n, d} > b} \cdot \ind{y_n = +1}) + \hat{N} & (\hatyl = +1 \land \hatyr = -1) \\
            \frac{1}{N} \sum_{n=1}^{N} (f_{n, i} \cdot \ind{x_{n, d} \leq b} \cdot \ind{y_n = +1} + f_{n, i} \cdot \ind{x_{n, d} > b} \cdot \ind{y_n = -1}) + \hat{N} & (\hatyl = -1 \land \hatyr = +1) \\
            \frac{1}{N} \sum_{n=1}^{N} f_{n, i} \cdot \ind{y_n = -1} + \hat{N} & (\hatyl = +1 \land \hatyr = +1)
        \end{cases} \\
        &=
        \begin{cases}
            \frac{N^+ + \bar{N}}{N} & (\hatyl = -1 \land \hatyr = -1) \\
            \frac{N^-_\mathrm{L} + N^+_\mathrm{R} + \bar{N}}{N} & (\hatyl = +1 \land \hatyr = -1) \\
            \frac{N^+_\mathrm{L} + N^-_\mathrm{R} + \bar{N}}{N} & (\hatyl = -1 \land \hatyr = +1) \\
            \frac{N^- + \bar{N}}{N} & (\hatyl = +1 \land \hatyr = +1),
        \end{cases}        
    \end{align*}
    which shows the correctness of \cref{eq:er}. 
    Thus, we can compute the empirical risk $\hat{R}(h' \mid S)$ correctly using the terms $\hat{N}, N^+, N^-, N^+_\mathrm{L}, N^-_\mathrm{L}, N^+_\mathrm{R}, N^-_\mathrm{R}$. 

    From the above results, \cref{algo:ract} can compute $\hat{R}(h' \mid S)$ and $\hat{\Omega}_\epsilon(h' \mid S)$ for a fixed $(d, b)$ and $\hatyl, \hatyr$. 
    Furthermore, we can solve the inner optimization problem of \cref{eq:split} by enumerating four patterns of $\hatyl, \hatyr \in \set{ \pm 1 }$. 
    Since \cref{algo:ract} searches all the pairs $(d, b)$ of a feature $d \in [D]$ and threshold $b \in B_d$ and computes the optimal value of the inner optimization problem of \cref{eq:split} for each $(d, b)$ in lines~7--27, it can solve the problem~\eqref{eq:split} in line~29, which concludes the proof of the correctness of \cref{algo:ract}. 

    \paragraph{Complexity.}
    Next, we prove that \cref{algo:ract} runs in $\mathcal{O}(D \cdot N)$. 
    Our complexity analysis of \cref{algo:ract} can be divided into the following five parts:
    \begin{itemize}
        \item 
        In lines~2--5, \cref{algo:ract} initializes the terms in $\mathcal{O}(N)$. 
        \item
        In the for-loop of lines~10--26, each while-loop runs at most $N$ times through the for-loop of a fixed $B_d$.  
        In addition, the optimization task in line~25 can be solved by enumerating four patterns of $\hatyl, \hatyr$, which can be regarded as a constant time. 
        Since we assume $|B_d| = \mathcal{O}(N)$, the overall complexity of lines~10--26 is $\mathcal{O}(N)$. 
        \item 
        Since the inner for-loop of lines~10--26 takes $\mathcal{O}(N)$, the outer for-loop of lines~7--27 takes $\mathcal{O}(D \cdot N)$. 
        \item 
        The optimization task in line~29 can be solved in $\mathcal{O}(D \cdot N)$ because the objective value of each $(d, b)$ has been computed in lines~7-27. 
        \item 
        In lines~31--35, \cref{algo:ract} updates the terms in $\mathcal{O}(N)$.         
    \end{itemize}
    In summary, the overall complexity of \cref{algo:ract} is $\mathcal{O}(D \cdot N)$, which concludes the proof. 
\end{proof}

\subsection{Proof of \cref{prop:wpc}}
We show that the problem~\eqref{eq:cover} is an instance of the weighted partial cover problem defined as follows~\cite{Chvatal:MOR1979,Kearns:1990:computational}. 

\begin{problem}[Weighted Partial Cover]\label{prob:appendix:wpc}
    For a positive integer $I, M \in \mathbb{N}$, let $S_1, \dots, S_I \subseteq [M]$ be $I$ subsets of $[M]$ with positive real weights $w_1, \dots, w_I \geq 0$, and we assume that $\bigcup_{i=1}^{I} S_i = [M]$. 
    Given $p \in (0, 1]$, find an optimal solution to the following problem:
\begin{align}
    \mathop{\text{minimize}}_{\mathcal{I} \subseteq [I]} \; \sum_{i \in \mathcal{I}} w_i \;\;\; \text{subject to} \;\; \frac{1}{M} \left| \bigcup_{i \in \mathcal{I}} S_i \right| \geq p. 
\end{align} 
\end{problem}

\begin{proof}[Proof of \cref{prop:wpc}]
    Let $\tilde{\mathcal{N}} = \{ n \in [N] \mid \exists i' \in \mathcal{I}^+, \exists \ba \in \calA_\varepsilon(\bx_n): \bx_n + \ba \in r_{i'} \}$ be the set of instances that have at least one reachable leaf $i'$ with $\haty_{i'} = +1$. 
    Using $\tilde{\mathcal{N}}$, we have $\left|{\bigcup}_{i \in \mathcal{I}^+ \cup \mathcal{I}} \mathcal{N}_i\right| = \tilde{N} + \left|{\bigcup}_{i \in \mathcal{I}} \tilde{\mathcal{N}}_i\right|$, where $\tilde{N} = |\tilde{\mathcal{N}}|$ and $\tilde{\mathcal{N}}_i = \{ n \in [N] \mid n \not\in \tilde{\mathcal{N}} \land \exists a \in \mathcal{A}_\varepsilon(x_n): x_n + a \in r_i \}$. 
    Then, we can express the constraint of \cref{eq:cover} as 
    \begin{align*}
        \frac{1}{N} \left|\bigcup_{i \in \mathcal{I}^+ \cup \mathcal{I}} \mathcal{N}_i\right| \geq 1 - \delta \iff \frac{1}{N} \left|\bigcup_{i \in \mathcal{I}} \tilde{\mathcal{N}}_i\right| \geq 1 - \delta - \frac{\tilde{N}}{N}.     
    \end{align*}
    Thus, we can see that the problem of \cref{eq:cover} is an instance of the weighted partial cover problem defined by \cref{prob:appendix:wpc} by replacing 
    (i)~$p$ with $1 - \delta - \frac{\tilde{N}}{N}$, 
    (ii)~$S_i$ with $\tilde{\mathcal{N}}_i$, 
    (iii)~$M$ with $N$, 
    (iv)~$w_i$ with $c_i$, and
    (v)~$[I]$ with $\mathcal{I}^-$, respectively. 
\end{proof}

\subsection{Proof of \cref{prop:pac}}
\begin{proof}[Proof of \cref{prop:pac}]
    Recall that our empirical recourse risk is defined as $\hat{\Omega}(h \mid S) = \frac{1}{N} \sum_{n=1}^{N} l_\mathrm{rec}(\bx_n; h)$. 
    Let $\mathcal{D}$ be a distribution over the input domain $\mathcal{X}$. 
    By definition of our recourse loss $l_\mathrm{rec}$, we have
    \begin{align*}
        \Omega(h) &= \mathop{\mathbb{P}}_{\bx \sim \mathcal{D}}[\exists \ba \in \calA_\epsilon(\bx): h(\bx+\ba) = +1] \\
        &= \mathop{\mathbb{E}}_{\bx \sim \mathcal{D}}[\mathbb{I}[\exists \ba \in \calA_\epsilon(\bx): h(\bx+\ba) = +1]] \\
        &= \mathop{\mathbb{E}}_{\bx \sim \mathcal{D}}[\min_{\ba \in \calA_\epsilon(\bx)} l_{01}(+1, h(\bx + \ba))] \\
        &= \mathop{\mathbb{E}}_{\bx \sim \mathcal{D}}[l_\mathrm{rec}(\bx; h)].
    \end{align*}
    It implies that $\Omega$ is the expected value of the recourse loss $l_\mathrm{rec}$ over the distribution $\mathcal{D}$ and that $\hat{\Omega}$ is the empirical average of $l_\mathrm{rec}$ over a finite set of $N$ instances $\bx_1, \dots, \bx_N$ when they are i.i.d. samples drown from $\mathcal{D}$. 

    Here, we consider the probability of $\Omega(h) - \hat{\Omega}(h \mid S) > \theta$ for some $\theta > 0$ and a fixed classifier $h \in \mathcal{H}$. 
    Recall that our recourse loss $l_\mathrm{rec}$ is bounded in $[0,1]$. 
    By applying the Hoeffding's inequality~\cite{Mohri:2012:Foundations}, we have
    \begin{align*}
        \mathbb{P}[\Omega(h) - \hat{\Omega}(h \mid S) > \theta] \leq \exp \left( -2 \cdot N \cdot \theta^2 \right). 
    \end{align*}
    Next, we consider bounding the probability that there exists $h \in \mathcal{H}$ such that $\Omega(h) - \hat{\Omega}(h \mid S) > \theta$ by $\alpha > 0$. 
    Using the union bound, we have
    \begin{align*}
        \mathbb{P}[\forall h \in \mathcal{H}: \Omega(h) - \hat{\Omega}(h \mid S) > \theta] 
        \leq \sum_{h \in \mathcal{H}} \mathbb{P}[\Omega(h) - \hat{\Omega}(h \mid S) > \theta]
        \leq |\mathcal{H}| \cdot \exp \left( -2 \cdot N \cdot \theta^2 \right). 
    \end{align*}
    Therefore, we obtain
    \begin{align*}
        |\mathcal{H}| \cdot \exp \left( -2 \cdot N \cdot \theta^2 \right) = \alpha \iff \theta = \sqrt{\frac{\ln |\mathcal{H}| - \ln \alpha}{2 N}}.
    \end{align*}
    In summary, the following inequality holds with probability $1 - \alpha$:
    \begin{align*}
        \Omega(h) - \hat{\Omega}(h \mid S) \leq \theta \iff \Omega(h) \leq \hat{\Omega}(h \mid S) + \sqrt{\frac{\ln |\mathcal{H}| - \ln \alpha}{2 N}},
    \end{align*}
    which concludes the proof. 
\end{proof}

\section{Implementation Details}
\subsection{Baseline Methods}
\paragraph{Vanilla.}
To the best of our knowledge, there is no study on learning tree-based models while guaranteeing the existence of recourse actions. 
Thus, as baseline approaches, we employed \emph{classification and regression trees~(CART)}~\cite{Breiman:1984:CART} and \emph{random forest}~\cite{Breiman:ML2001}, which are standard learning frameworks for classification trees and tree ensembles. 

\paragraph{Only Actionable Features (OAF). }
As another baseline, we employed a modified version of CART and random forest, named \emph{only actionable features~(OAF)}, that uses only features that can be changed by actions. 
Specifically, this approach trains classification trees and tree ensembles using only features that are not specified as immutable in each dataset. 
This idea is based on the observation of the existing study that relying on actionable features facilitates the existence of recourse actions~\cite{Dominguez-olmedo:ICML2022}. 
Our setting of immutable features in each dataset is shown in \cref{sec:appendix:additional}.

\subsection{Algorithms for Extracting Actions from Tree-Based Models}
\begin{algorithm}[t]
    \caption{Actionable feature tweaking algorithm for extracting actions from tree-based models~\cite{Tolomei:KDD2017}. }
    \small
    \begin{algorithmic}[1]
        \REQUIRE{
            a tree ensemble classifier $h$ with $T$ classification trees $h_1, \dots, h_T$,
            the region $r_{t, i}$ and predictive label $\haty_{t, i}$ of each leaf $i \in [I_t]$ in each tree $h_t$,
            and an instance $\bx$ such that $h(\bx) \not= +1$. 
        }
        \ENSURE{an action $\ba^\ast$. }
        \STATE $\ba^\ast \leftarrow 0$; $c^\ast \leftarrow \infty$; 
        \STATE /* Enumerate All Leaves in Tree Ensemble */
        \FOR{$t = 1, \dots, T$}
            \FOR{$i = 1, \dots, I_t$}
                \STATE /* Find Leaf with Desired Class Label */
                \IF{$\haty_{t, i} = y^\ast$}
                    \STATE /* Optimize Action for Leaf */
                    \STATE $\hat{\ba} \leftarrow \arg\min_{\ba \in \calA(\bx)} c(\ba \mid \bx)$ s.t. $\bx + \ba \in r_{t, i}$; 
                    \IF{$h(\bx + \hat{\ba}) = +1$ and $c(\hat{\ba} \mid x) < c^\ast$}
                        \STATE $\ba^\ast \leftarrow \hat{\ba}$; $c^\ast \leftarrow c(\hat{\ba} \mid \bx)$; 
                    \ENDIF
                \ENDIF
            \ENDFOR
        \ENDFOR
        \STATE \textbf{return} $\ba^\ast$; 
    \end{algorithmic}
    \label{algo:appendix:tweaking}
\end{algorithm}

To extract actions from tree-based models, there exist algorithms based on several mathematical techniques, such as integer optimization~\cite{Cui:KDD2015,Kanamori:IJCAI2020,Parmentier:ICML2021}, probabilistic approximation~\cite{Lucic:AAAI2022}, and heuristic methods~\cite{Tolomei:KDD2017,Carreira:AAAI2023}. 
In this paper, we employed \emph{actionable feature tweaking algorithm}~\cite{Tolomei:KDD2017} which is a fast heuristic method. 
The reason why we used this method is that it can handle additional constraints, such as plausibility~\cite{Parmentier:ICML2021} and causality~\cite{Karimi:FAccT2021}. 

\cref{algo:appendix:tweaking} presents a pseudo-code of the actionable feature tweaking algorithm. 
\cref{algo:appendix:tweaking} consists of the following three steps:
(i)~for each classification tree $h_t$ in the ensemble $h$, enumerating all the leaves $i$ whose predictive label is the desired class (i.e., $\haty_{t, i} = +1$); 
(ii)~computing an optimal action $\hat{\ba}$ to the region $r_{t, i}$ corresponding to the leaf $i$; 
(iii)~finding the minimum cost action $\ba^\ast$ among ones altering the prediction results of $h$ into the desired class (i.e., $h(\bx + \ba) = +1$). 
Note that we can easily compute an optimal action $\hat{\ba} = (\hat{a}_1, \dots, \hat{a}_D)$ to an instance $\bx$ and a region $r = [l_1, u_1) \times \dots \times [l_D, u_D)$ as $\hat{a}_d = \operatorname{median}(x_d, l_d, u_d) - x_d$ for $d \in [D]$~\cite{Wang:ICML2020,Carreira:AAAI2023}. 
In addition, our implementation is parallelized using Numba\footnote{\url{https://numba.pydata.org/}} and runs faster than the existing public implementations.

\subsection{Greedy Algorithm for Set-Cover Relabeling}
\begin{algorithm}[t]
    \caption{Greedy approximation algorithm for the weighted partial cover problem~\cite{Kearns:1990:computational}. }
    \small
    \begin{algorithmic}[1]
        \REQUIRE{
            sets of leaves $\mathcal{I}^+, \mathcal{I}^- \subseteq [I]$ whose predictive labels are $+1$ and $-1$,  
            a set of instances $\mathcal{N}_i$ that can reach the leaf $i$ by some action for each $i \in [I]$, 
            weight values $c_i = N^-_i - N^+_i$ for each $i \in [I]$, 
            and a parameter $\delta \geq 0$. 
        }
        \ENSURE{A set $\mathcal{I}$ of leaves that should be modified as $\haty_i = +1$. }
        \STATE Define $g(\mathcal{I}) \coloneqq \frac{1}{N} \left|\bigcup_{i \in \mathcal{I}^+ \cup \mathcal{I}} \mathcal{N}_i\right|$; 
        \STATE $\mathcal{I} \leftarrow \emptyset$; 
        \STATE /* Greedy Optimization */
        \WHILE{$g(\mathcal{I}) < 1 - \delta$}
            \STATE $\hat{i} \leftarrow \arg\max_{i \in \mathcal{I}^-} \frac{g(\mathcal{I} \cup \set{i}) - g(\mathcal{I})}{c_i}$; 
            \STATE $\mathcal{I} \leftarrow \mathcal{I} \cup \set{ \hat{i} }$; 
            \STATE $\mathcal{I}^- \leftarrow \mathcal{I}^- \setminus \set{\hat{i}}$; 
        \ENDWHILE
        \STATE \textbf{return} $\mathcal{I}$; 
    \end{algorithmic}
    \label{algo:appendix:greedy}
\end{algorithm}

While the problem~\eqref{eq:cover} is NP-hard because it is an instance of the weighted partial cover problem, there exist several polynomial time algorithms with approximation guarantees~\cite{Chvatal:MOR1979}. 
\cref{algo:appendix:greedy} presents a greedy approximation algorithm for the problem. 
It has a $(2 \cdot H(N) + 3)$-approximation guarantee, where $H(N) = \sum_{n=1}^{N} 1/n = \Theta(\ln N)$~\cite{Kearns:1990:computational}.

\section{Detailed Experimental Settings and Additional Results}\label{appendix:exp}
\begin{table}[t]
    \centering
      \caption{
        Details of hyper-parameter tuning for our RACT. 
        We varied the values of $\delta$ and $\lambda$ in the predefined ranges (\textbf{Range}), and determined each value (\textbf{Select}) based on the results. 
    }
\vskip 0.1in
    \label{tab:appendix:exp:parameter}
    \subtable[Classification Tree]{
        \begin{tabular}{ccccc}
        \toprule
            & \multicolumn{2}{c}{$\delta$} & \multicolumn{2}{c}{$\lambda$} \\
            \cmidrule(lr){2-3} \cmidrule(lr){4-5}
            \textbf{Dataset} & \textbf{Range} & \textbf{Select} & \textbf{Range} & \textbf{Select} \\
        \midrule
            FICO   & $\set{0.2, 0.25, 0.3, 0.35, 0.4}$ & $0.3$ & $\set{0.025, 0.05, 0.075, 0.1}$ & $0.1$ \\
            COMPAS & $\set{0.2, 0.25, 0.3, 0.35, 0.4}$ & $0.3$ & $\set{0.025, 0.05, 0.075, 0.1}$ & $0.05$ \\
            Credit & $\set{0.2, 0.25, 0.3, 0.35, 0.4}$ & $0.3$ & $\set{0.025, 0.05, 0.075, 0.1}$ & $0.1$ \\
            Bail   & $\set{0.2, 0.25, 0.3, 0.35, 0.4}$ & $0.3$ & $\set{0.025, 0.05, 0.075, 0.1}$ & $0.05$ \\
        \bottomrule
        \end{tabular}
    }
    \hfill
    \subtable[Random Forest]{
        \begin{tabular}{ccc}
        \toprule
            & \multicolumn{2}{c}{$\lambda$} \\
            \cmidrule(lr){2-3}
            \textbf{Dataset} & \textbf{Range} & \textbf{Select} \\
        \midrule
            FICO   & $\set{0.2, 0.4, 0.6, 0.8, 1.0}$ & $1.0$ \\
            COMPAS & $\set{0.02, 0.04, 0.06, 0.08, 0.1}$ & $0.06$ \\
            Credit & $\set{0.2, 0.4, 0.6, 0.8, 1.0}$ & $0.2$ \\
            Bail   & $\set{0.02, 0.04, 0.06, 0.08, 0.1}$ & $0.1$ \\
        \bottomrule
        \end{tabular}
    }
\end{table}

\subsection{Complete Results}\label{sec:appendix:additional}
\cref{tab:appendix:exp:parameter} shows how we determined the hyper-parameters $\delta$ and $\lambda$ of our RACT. 
\cref{tab:appendix:datasets:fico,tab:appendix:datasets:compas,tab:appendix:datasets:credit,tab:appendix:datasets:bail} present the details on the value type, minimum value, maximum value, immutability, and constraint of each feature of the datasets that we used in our experiments. 

\subsubsection{Performance Comparison (\cref{sec:exp:comp})}
\cref{tab:appendix:exp:comp:time} presents the average running time of each method in 10-fold cross validation. 
For both classification trees and random forests, we can see that there is no significant difference in running time between the baselines and RACT on almost all the datasets. 

\begin{figure*}[p]
    \centering
    \subfigure[FICO]{
        \includegraphics[width=0.475\linewidth]{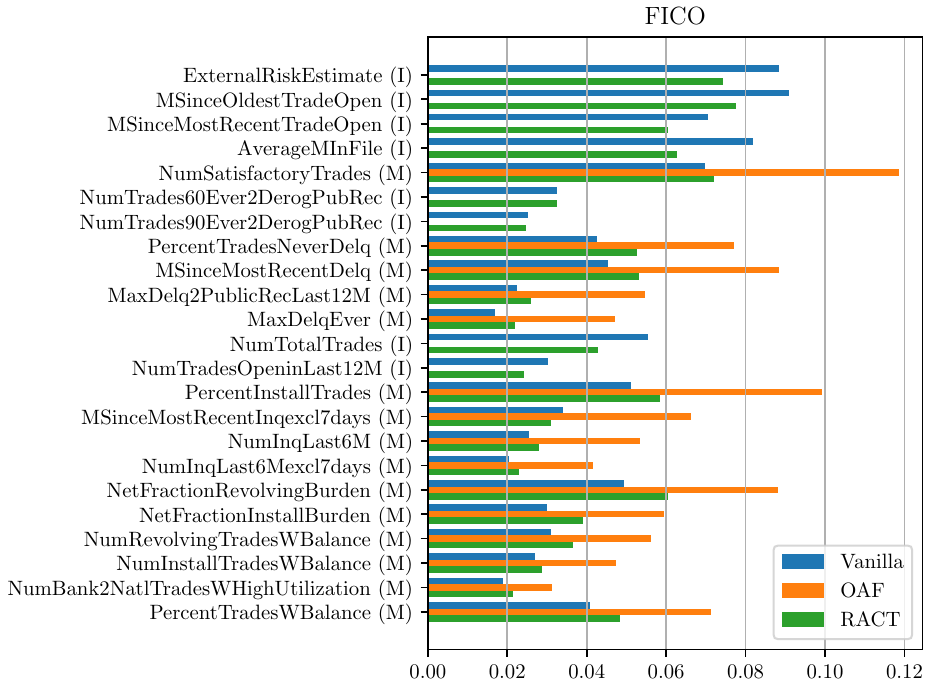}
    }
    \hfill
    \subfigure[COMPAS]{
        \includegraphics[width=0.475\linewidth]{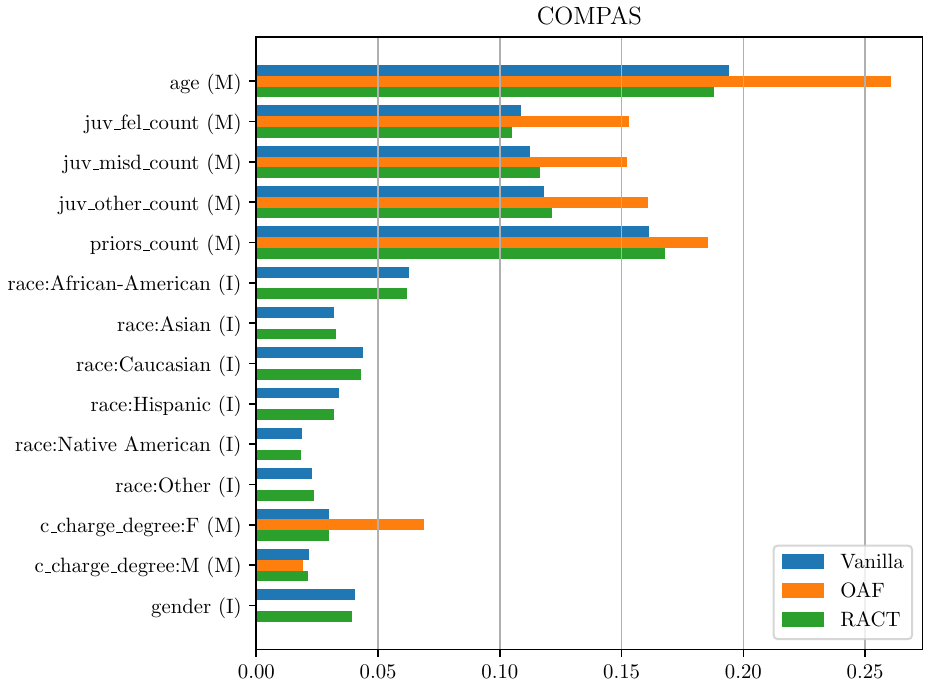}
    }
    \hfill
    \subfigure[Credit]{
        \includegraphics[width=0.475\linewidth]{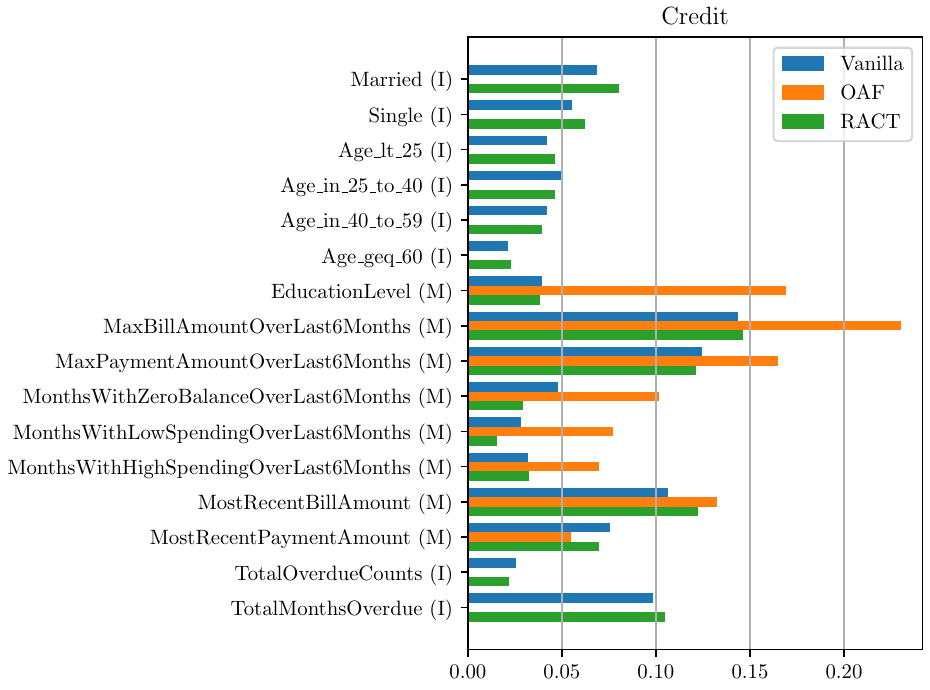}
    }
    \hfill
    \subfigure[Bail]{
        \includegraphics[width=0.475\linewidth]{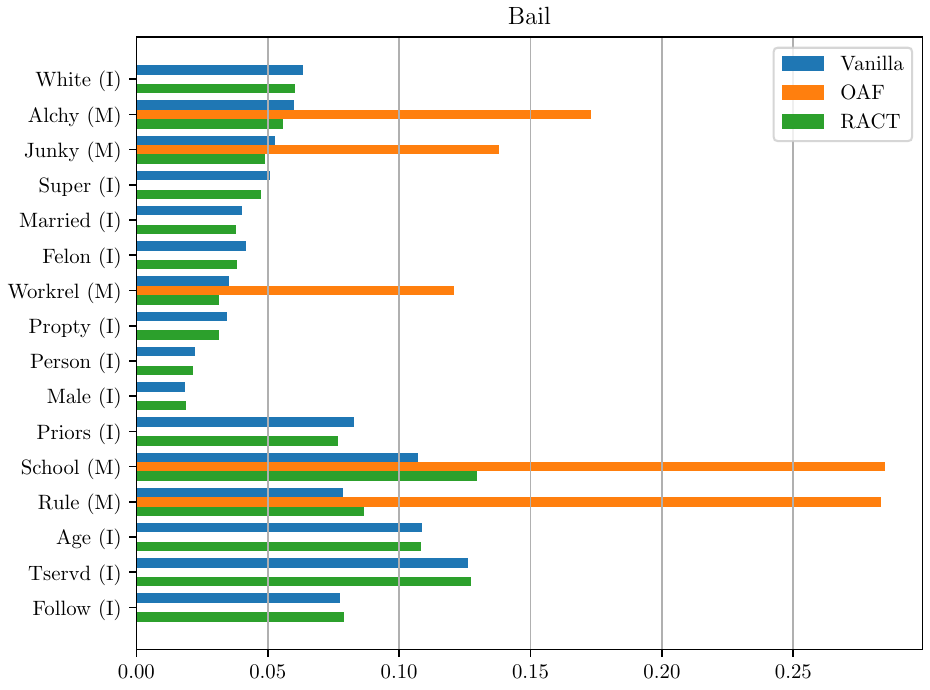}
    }
    \caption{
        Average feature importance of random forests leaned by each method in the performance comparison. 
        We measured the importance score of each feature by averaging the number of times the feature is used among trees, and normalized the scores so that the sum equals to $1$. 
        For each feature, ``I" (resp.\ ``M") stands for an immutable (resp.\ mutable) feature. 
    }
    \label{fig:appendix:exp:comp:importance}
\end{figure*}

\cref{fig:appendix:exp:comp:importance} shows the average feature importance of random forests leaned by each method in the performance comparison. 
We measured the importance score of each feature by averaging the number of times the feature is used among trees, as with the ``split" importance score of LightGBM~\cite{Ke:NIPS2017}, and normalized the scores so that the sum equals to $1$.

\subsubsection{Recourse Quality Analysis (\cref{sec:exp:qual})}
\cref{tab:appendix:exp:qual:cost,tab:appendix:exp:qual:plausibility,tab:appendix:exp:qual:causality} show the average cost, the average plausibility, and the average validity under the causal recourse constraint, respectively. 
For both classification trees and random forests, we observed that our RACT attained 
(i)~lower costs than the baselines; 
(ii)~comparable plausibility to the baselines; 
(iii)~higher validity than the baselines even in the causal recourse setting.

\subsubsection{Trade-Off Analysis (\cref{sec:exp:tradeoff})}
\begin{figure*}[p]
    \centering
    \includegraphics[width=\linewidth]{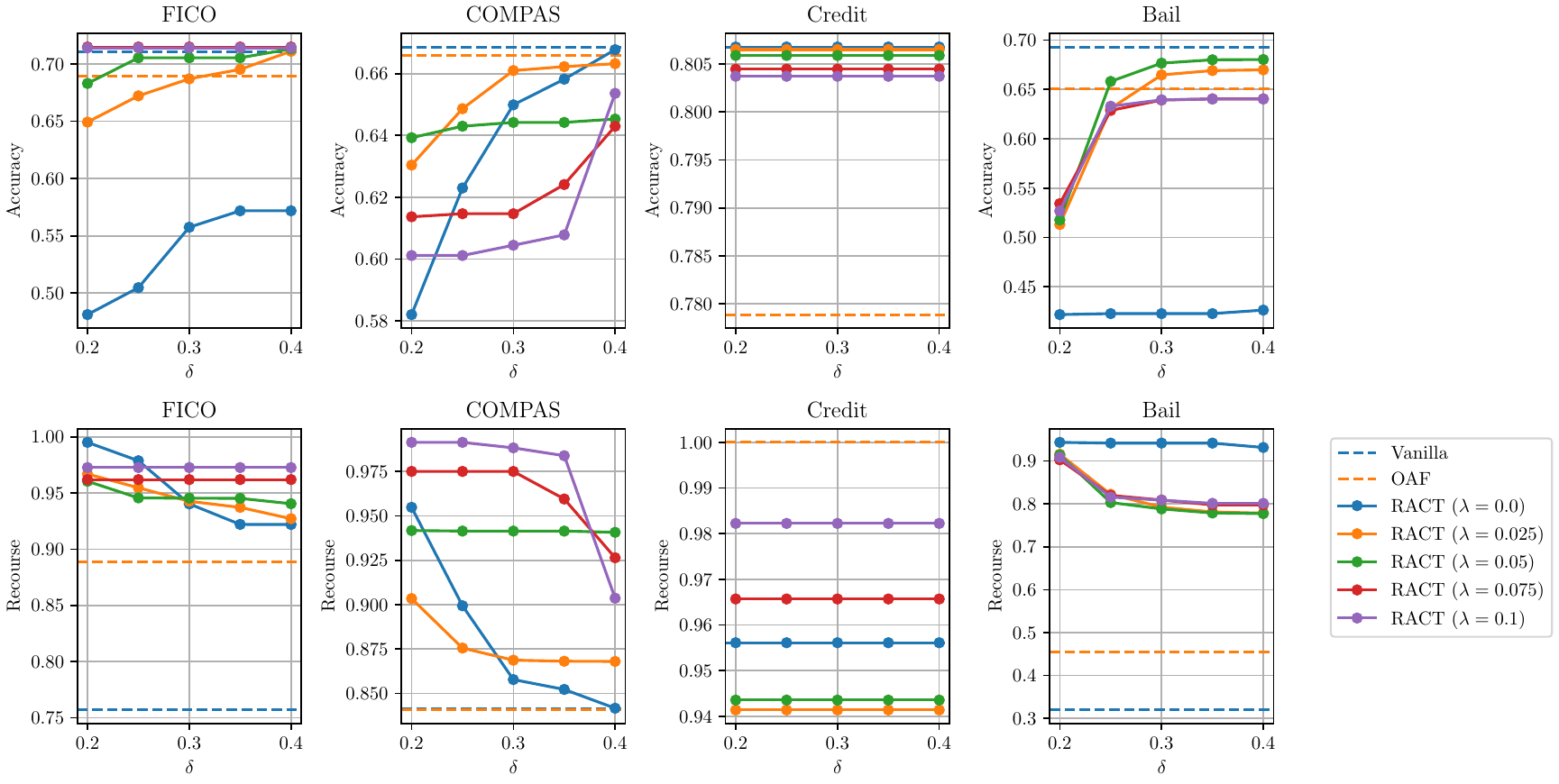}
    \caption{
        Sensitivity analyses of the parameter $\delta$ with respect to the average accuracy and recourse ratio of classification trees. 
        We can see the trend that the recourse ratio (resp.\ AUC) was improved by increasing (resp.\ decreasing) $\delta$. 
    }
    \label{fig:appendix:exp:tradeoff}
\end{figure*}

\cref{fig:appendix:exp:tradeoff} presents the sensitivity analyses of the parameter $\delta$ for classification trees. 
For each $\lambda \in \set{0.0, 0.025, 0.05, 0.075, 0.1}$, we measured the average accuracy and recourse ratio by varying $\delta \in \set{0.2, 0.25, 0.3, 0.35, 0.4}$. 
As with the results of random forests in the main paper, we can see that the recourse ratio (resp.\ accuracy) was improved by increasing (resp.\ decreasing) $\delta$ on almost all the datasets. 
We also observed that, for the FICO and Bail datasets, the average accuracy of $\lambda = 0.0$ was significantly worse than the others. 
This result suggests that applying our set-cover relabeling to a classification tree trained without our recourse risk $\hat{\Omega}_\epsilon$ often results in degrading accuracy, and that our set-cover relabeling can keep accuracy by combining our greedy splitting algorithm with $\hat{\Omega}_\epsilon$.

\subsection{Sensitivity Analysis of Model Complexity $T$}
\begin{figure*}[p]
    \centering
    \subfigure[$T=100$]{
        \includegraphics[width=\linewidth]{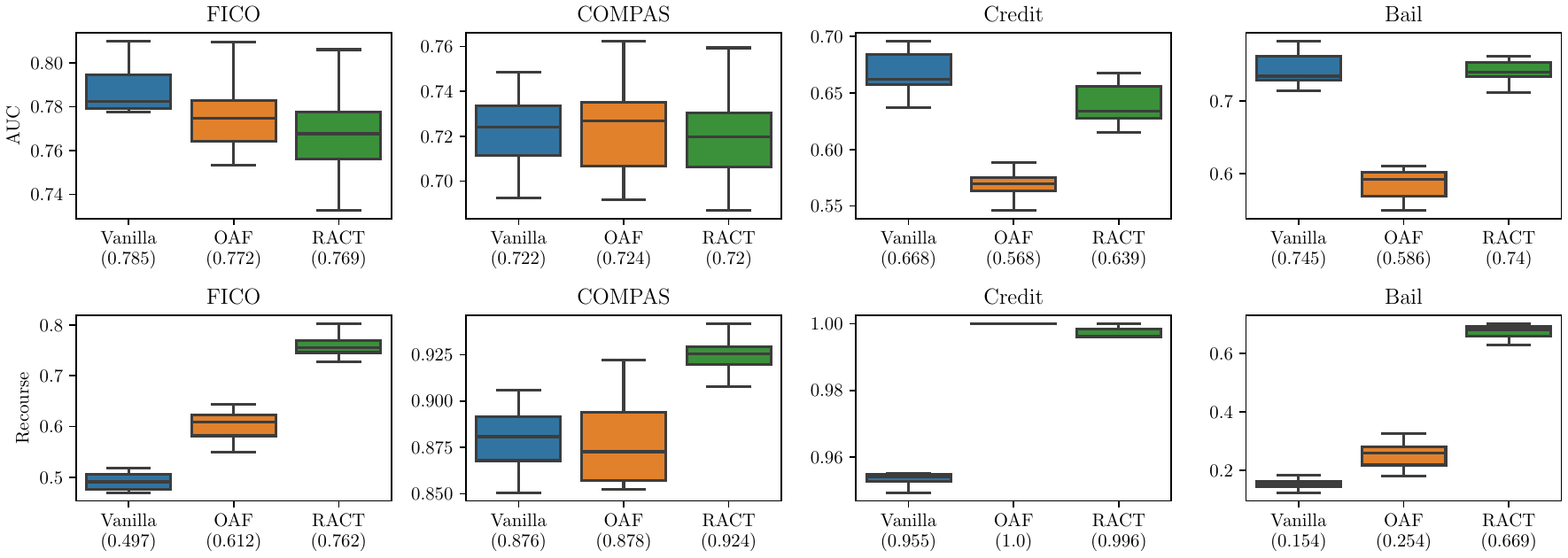}
    }
    \subfigure[$T=200$]{
        \includegraphics[width=\linewidth]{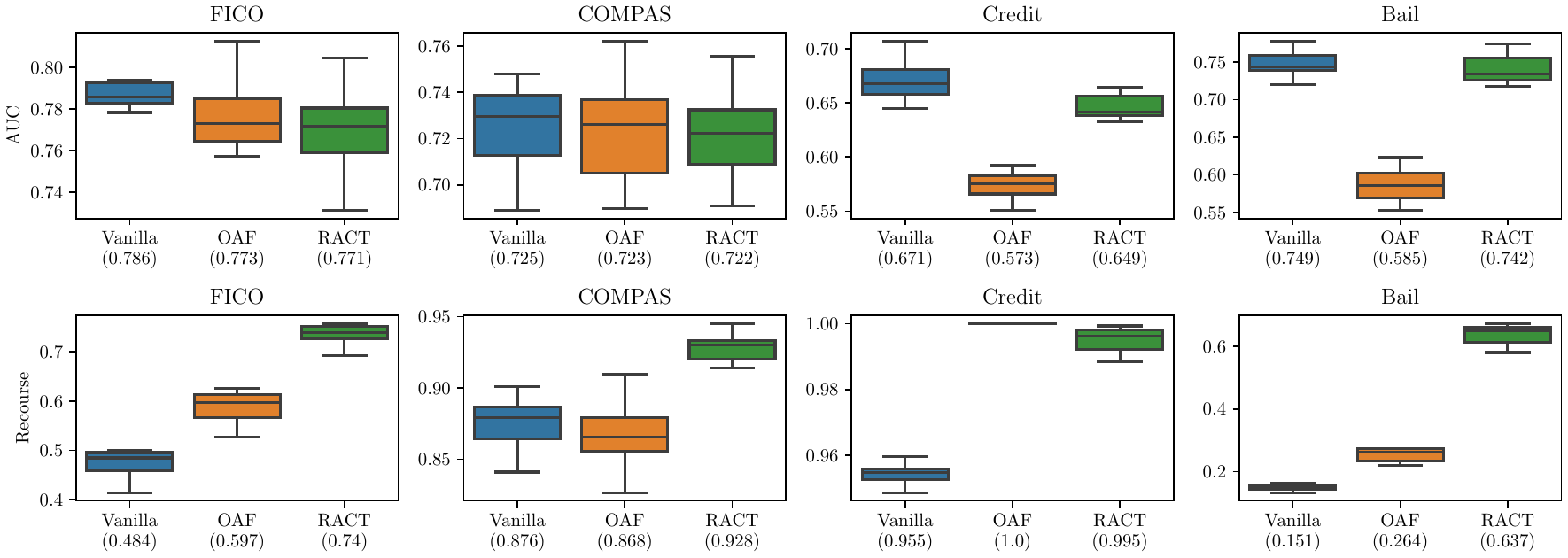}
    }
    \subfigure[$T=400$]{
        \includegraphics[width=\linewidth]{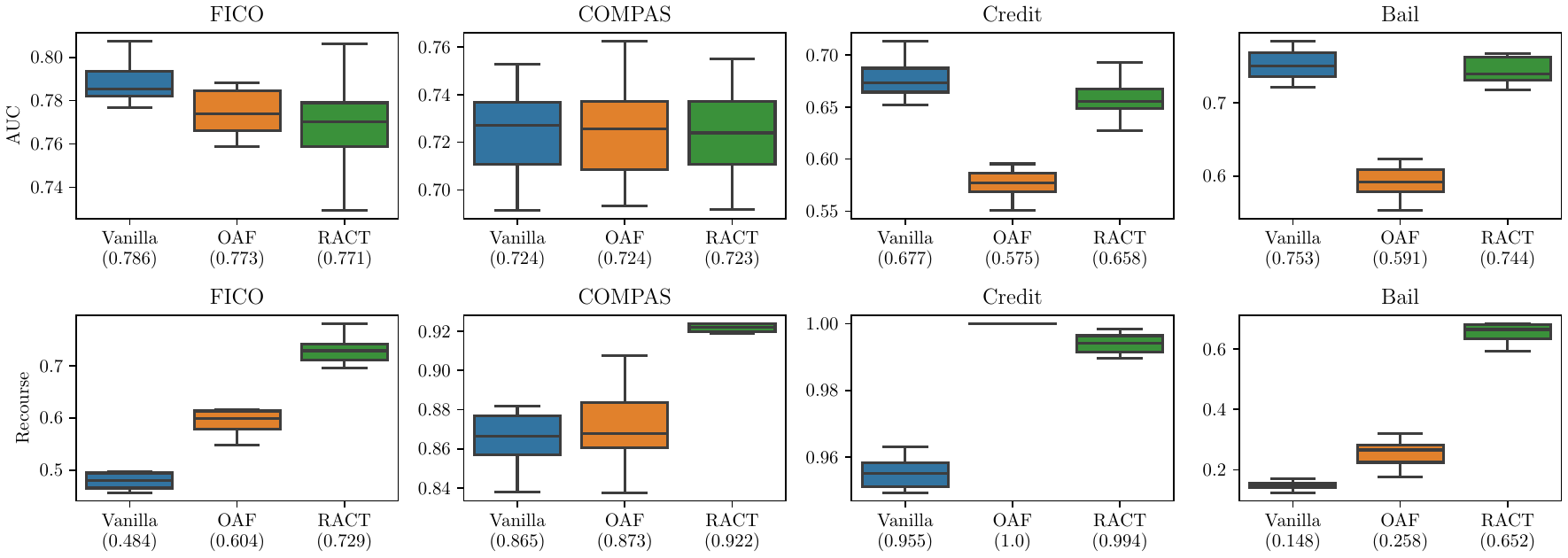}
    }
    \caption{
        Experimental results of the performance comparison for random forests with different numbers of trees $T \in \set{100, 200, 400}$. 
    }
    \label{fig:appendix:exp:complexity:comp}
\end{figure*}

\begin{figure*}[p]
    \centering
    \subfigure[$T=100$]{
        \includegraphics[width=\linewidth]{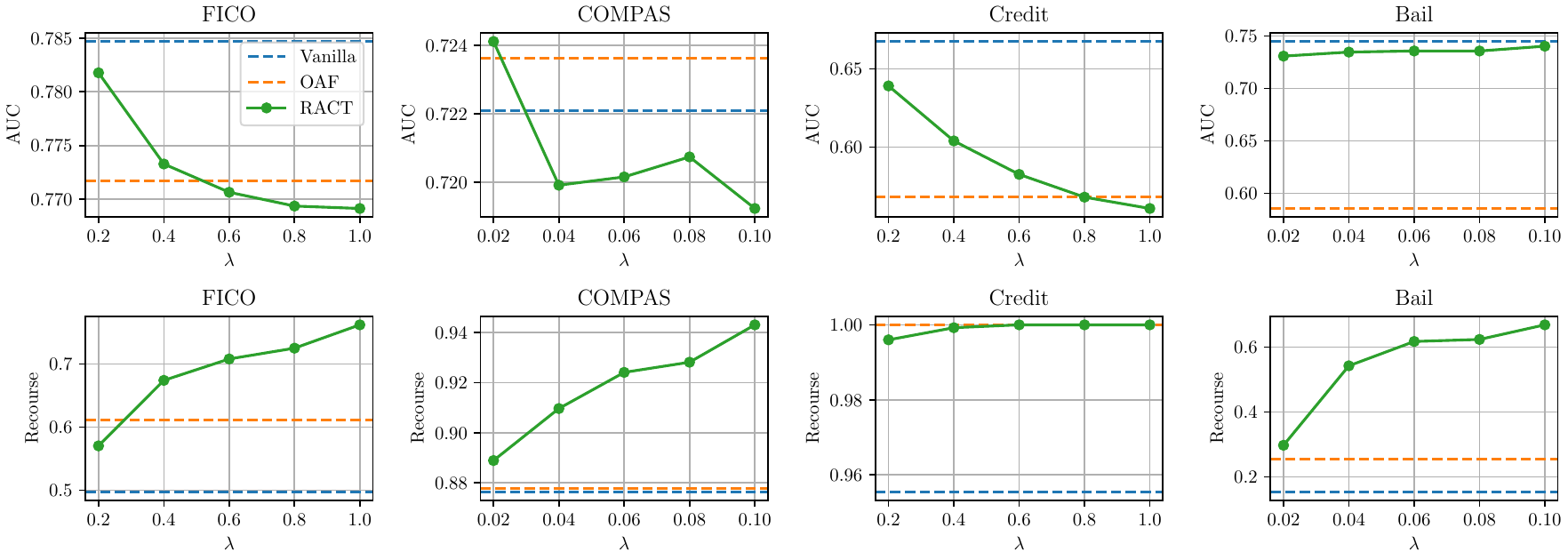}
    }
    \subfigure[$T=200$]{
        \includegraphics[width=\linewidth]{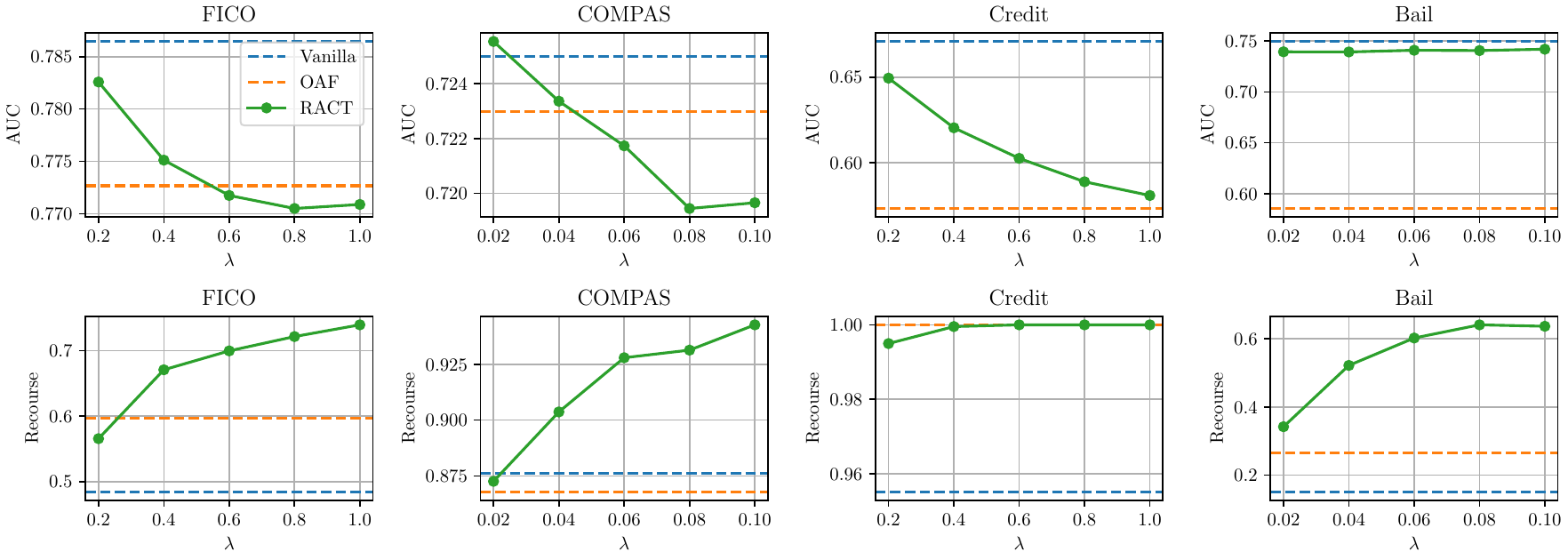}
    }
    \subfigure[$T=400$]{
        \includegraphics[width=\linewidth]{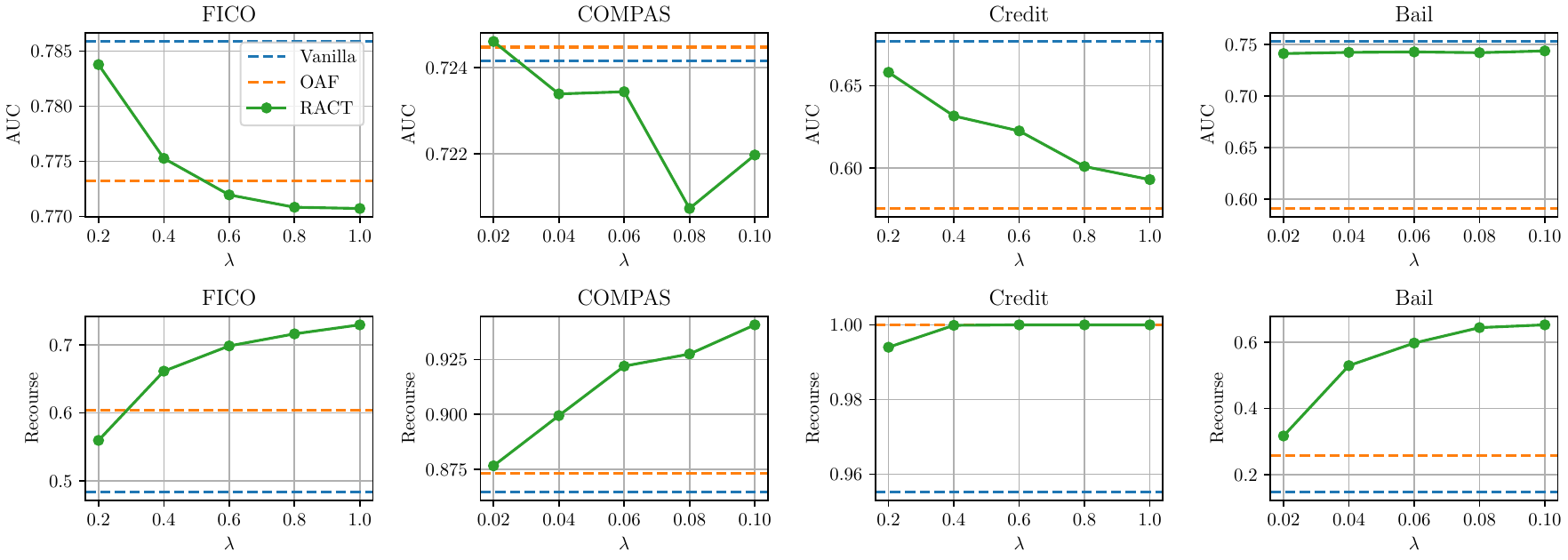}
    }
    \caption{
        Sensitivity analyses of the trade-off parameter $\lambda$ with respect to the average accuracy and recourse ratio of random forests for each number of trees $T \in \set{100, 200, 400}$. 
    }
    \label{fig:appendix:exp:complexity:tradeoff}
\end{figure*}

\cref{fig:appendix:exp:complexity:comp} presents the experimental results of our performance comparison with different model complexities for random forests. 
For each $T \in \set{100, 200, 400}$, we measured the average AUC and recourse ratio of each method, where we used the same value for $\lambda$ with \cref{sec:exp:comp}. 
\cref{fig:appendix:exp:complexity:tradeoff} presents the sensitivity analyses of the trade-off parameter $\lambda$ with different model complexities for random forests. 
As with \cref{sec:exp:tradeoff}, we measured the average AUC and recourse ratio of each method by varying $\lambda$.

\subsection{Sensitivity Analysis of Cost Budget $\epsilon$}
\begin{figure*}[p]
    \centering
    \subfigure[$\lambda=0.025$]{
        \includegraphics[width=0.9\linewidth]{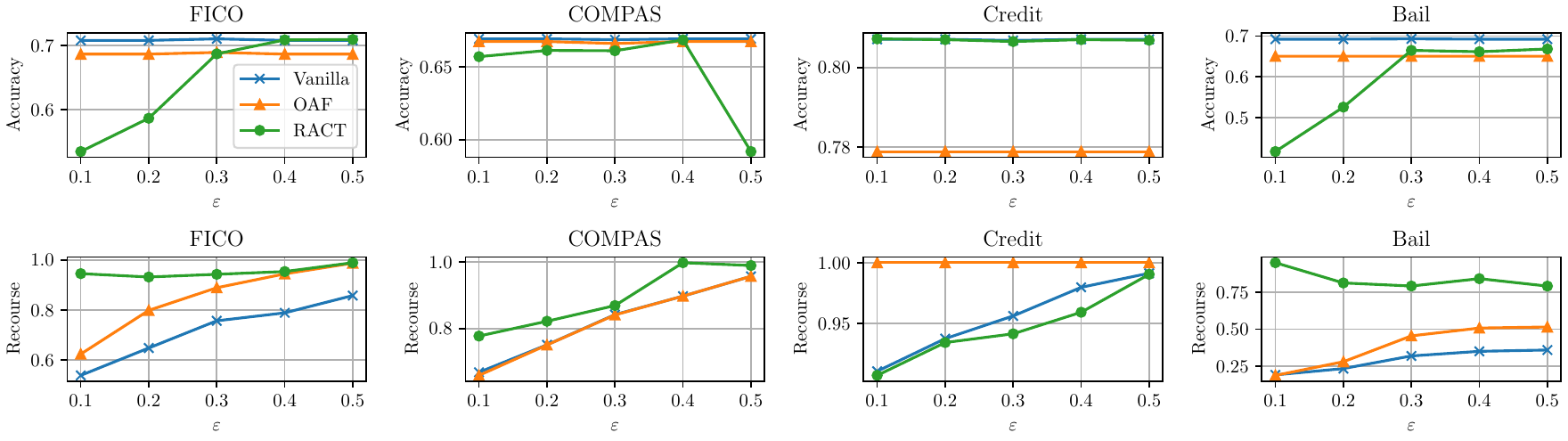}
    }
    \subfigure[$\lambda=0.05$]{
        \includegraphics[width=0.9\linewidth]{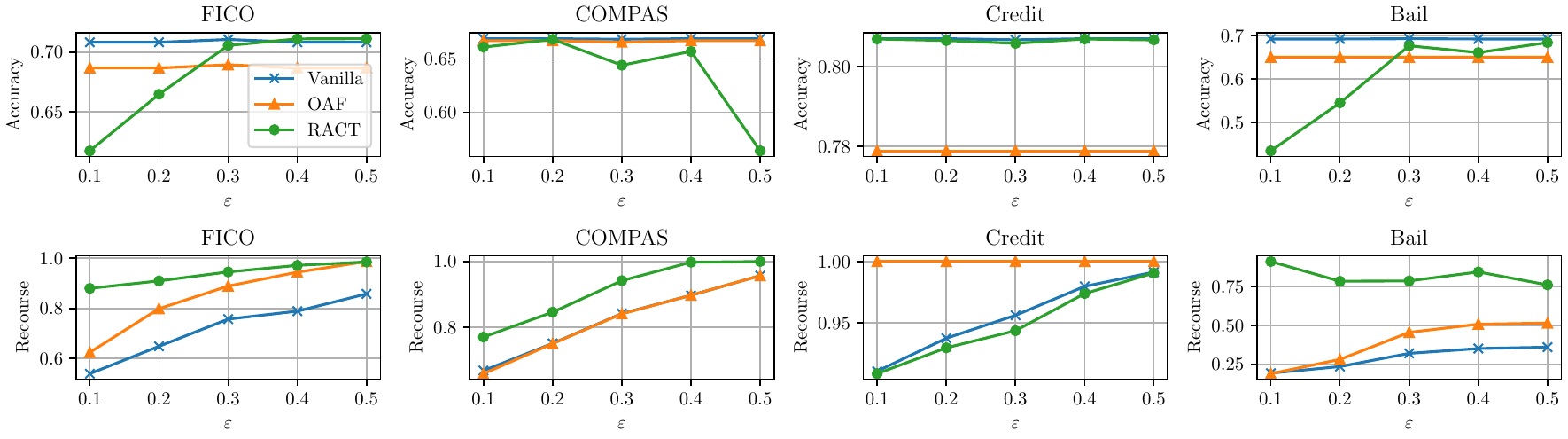}
    }
    \subfigure[$\lambda=0.075$]{
        \includegraphics[width=0.9\linewidth]{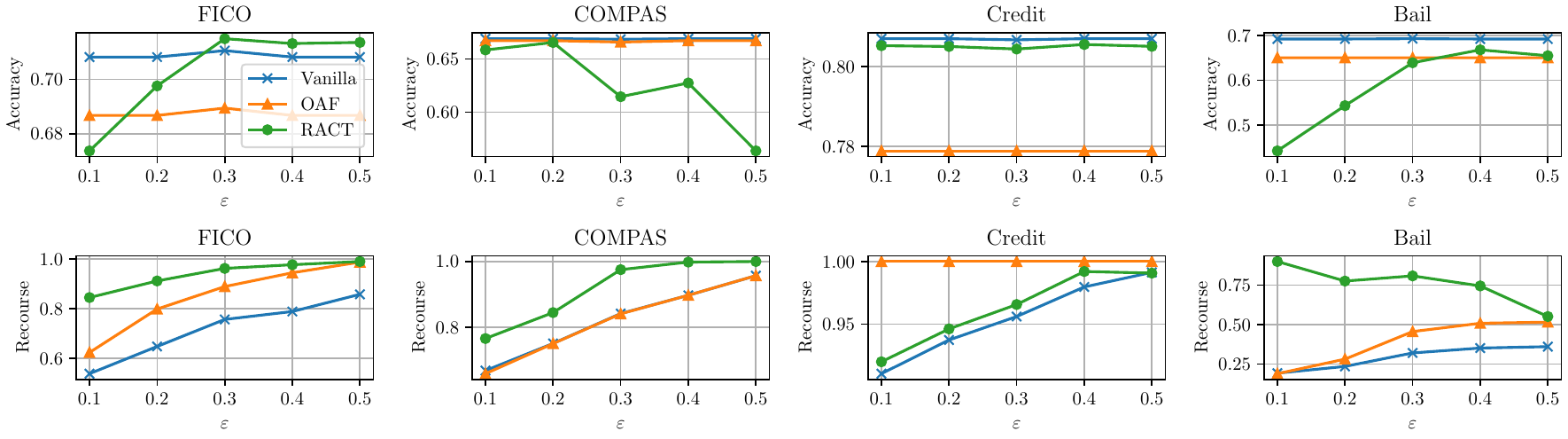}
    }
    \subfigure[$\lambda=0.1$]{
        \includegraphics[width=0.9\linewidth]{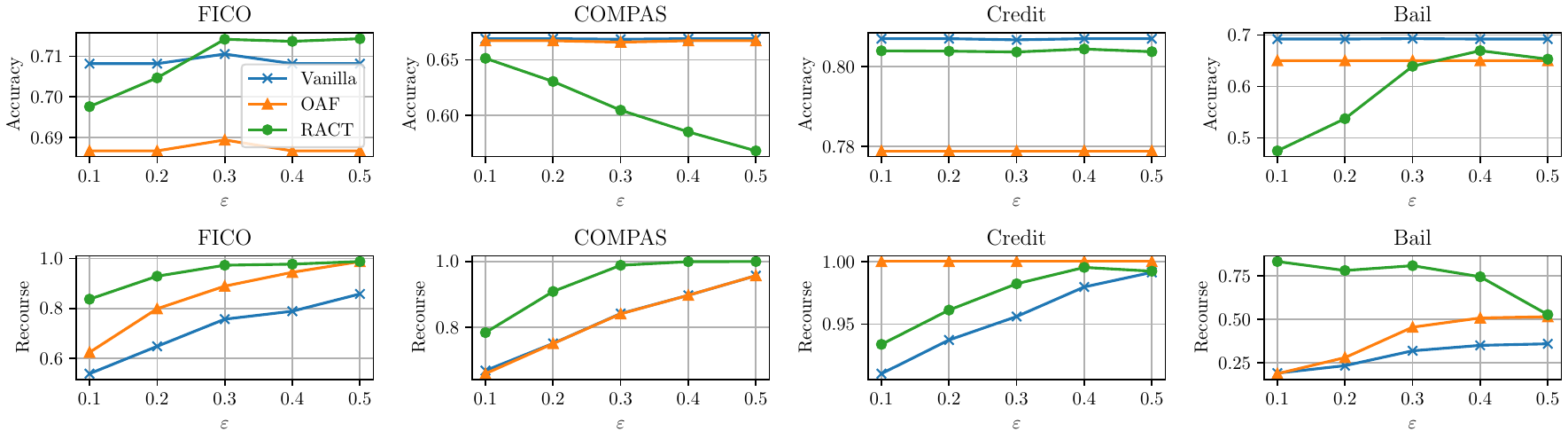}
    }
    \caption{
        Sensitivity analyses of the budget parameter $\epsilon$ with respect to the average accuracy and recourse ratio of classification trees. 
    }
    \label{fig:appendix:exp:sens:tree}
\end{figure*}

\begin{figure*}[p]
    \centering
    \subfigure[$\lambda=0.2$ for FICO and Credit, $\lambda = 0.02$ for COMPAS and Bail]{
        \includegraphics[width=0.725\linewidth]{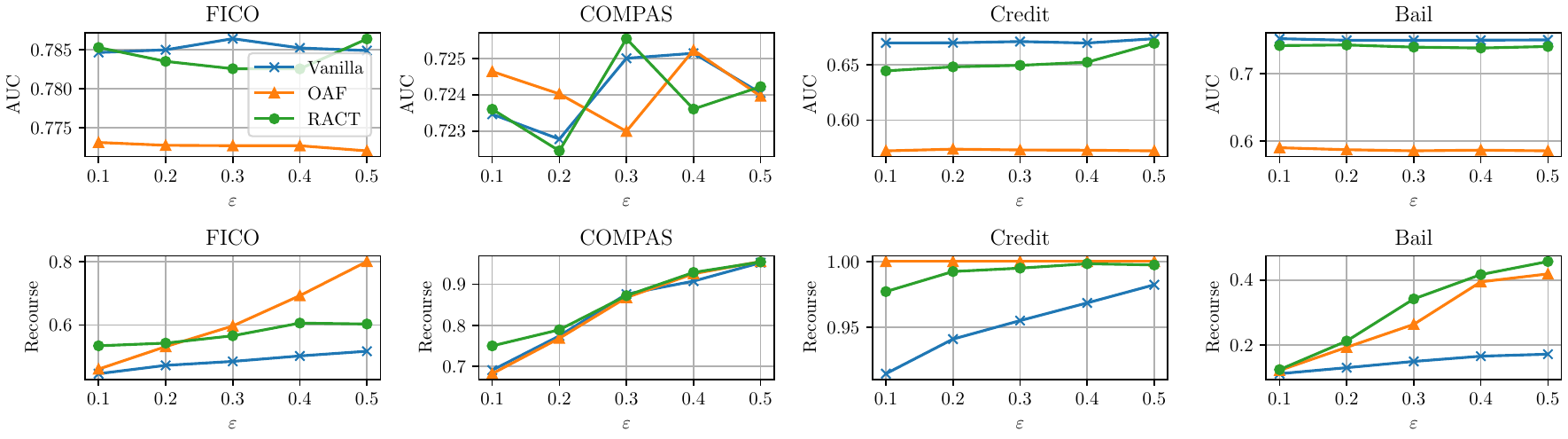}
    }
    \subfigure[$\lambda=0.4$ for FICO and Credit, $\lambda = 0.04$ for COMPAS and Bail]{
        \includegraphics[width=0.725\linewidth]{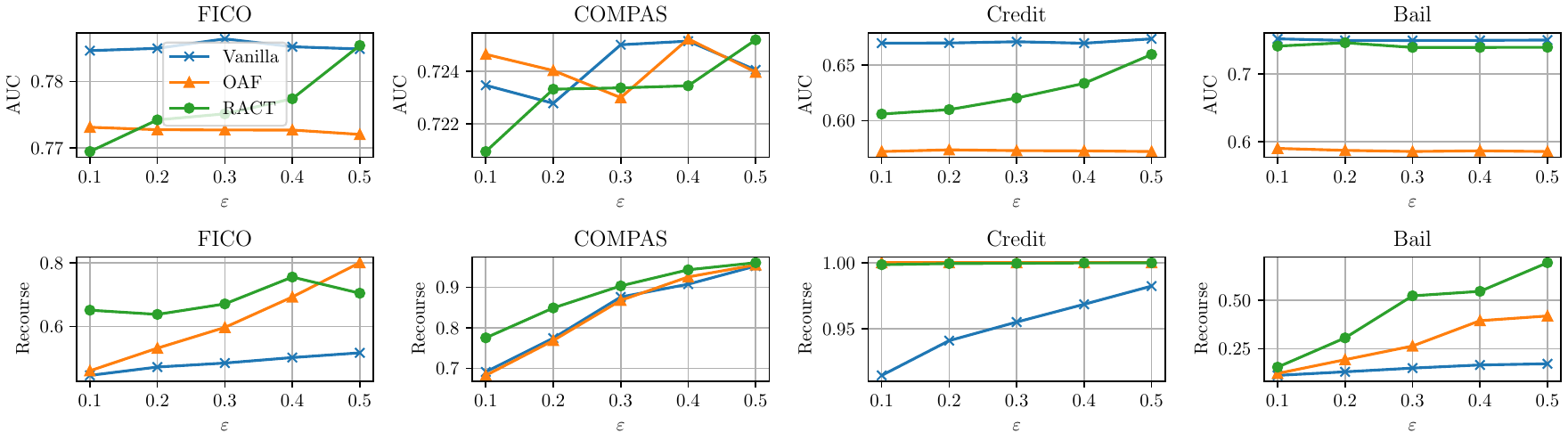}
    }
    \subfigure[$\lambda=0.6$ for FICO and Credit, $\lambda = 0.06$ for COMPAS and Bail]{
        \includegraphics[width=0.725\linewidth]{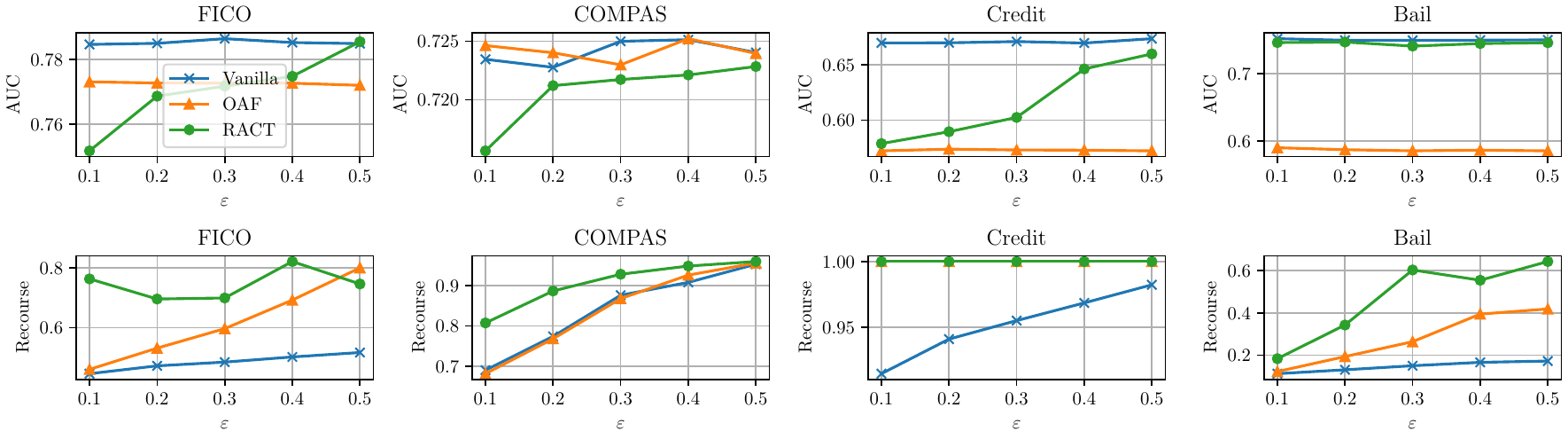}
    }
    \subfigure[$\lambda=0.8$ for FICO and Credit, $\lambda = 0.08$ for COMPAS and Bail]{
        \includegraphics[width=0.725\linewidth]{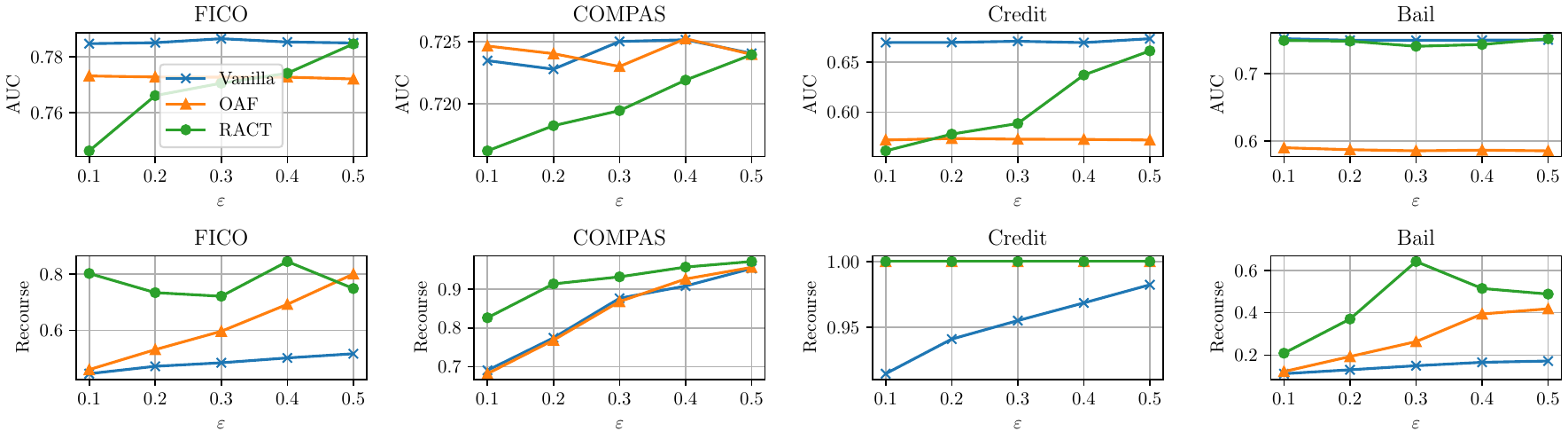}
    }
    \subfigure[$\lambda=1.0$ for FICO and Credit, $\lambda = 0.1$ for COMPAS and Bail]{
        \includegraphics[width=0.725\linewidth]{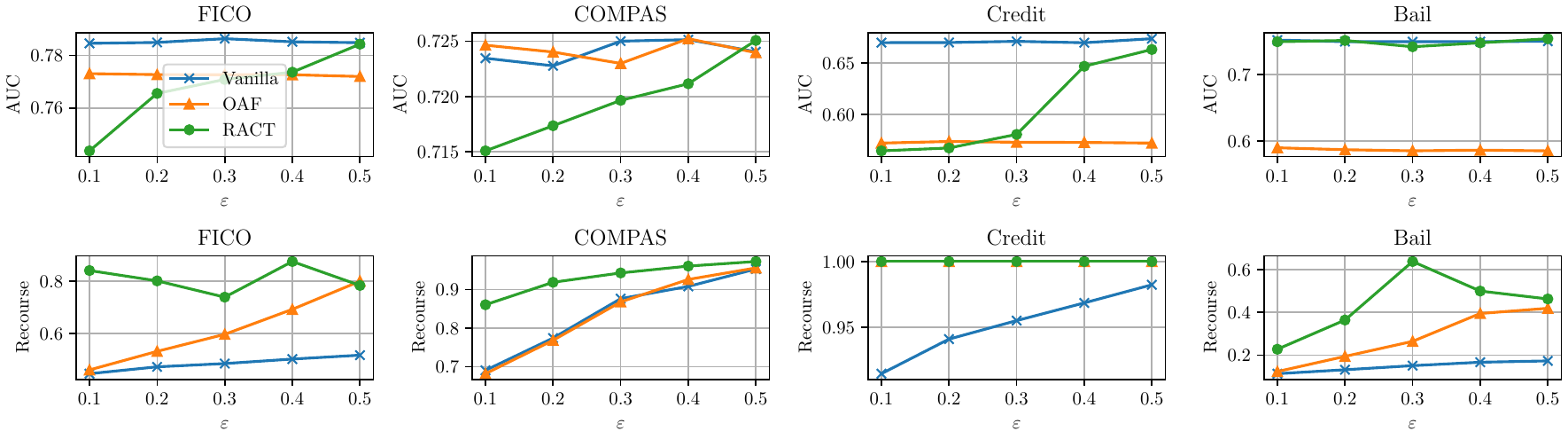}
    }
    \caption{
        Sensitivity analyses of the budget parameter $\epsilon$ with respect to the average accuracy and recourse ratio of random forests. 
    }
    \label{fig:appendix:exp:sens:forest}
\end{figure*}

\cref{fig:appendix:exp:sens:tree,fig:appendix:exp:sens:forest} present the sensitivity analyses of the budget parameter $\epsilon$ for classification trees and random forests, respectively. 
For each $\lambda$, we measured the average predictive performance and recourse ratio by varying $\epsilon \in \set{0.1, 0.2, 0.3, 0.4, 0.5}$.

\subsection{Comparison using Exact AR Algorithm based on Integer Optimization}
To investigate whether an action extraction algorithm affects the results, we also employed the exact AR method based on integer optimization~\cite{Cui:KDD2015,Kanamori:IJCAI2020} to extract actions from tree ensembles trained by each method. 
As with our experiment of \cref{sec:exp:comp}, we conducted 10-fold cross-validation and extracted actions for test instances predicted as the undesired class in each fold. 
As an off-the-shelf solver, we used Gurobi 10.0.3\footnote{\url{https://www.gurobi.com/}}, which is one of the state-of-the-art commercial solvers. 
Due to computational cost, we randomly picked $50$ test instances in each fold, and a $60$ second time limit was imposed on optimizing an action for each instance. 

\cref{tab:appendix:exp:milo:valid,tab:appendix:exp:milo:cost} presents the average validity and cost of the extracted actions. 
As with our results using the heuristic algorithm shown in \cref{sec:exp:comp}, we can see that our RACT achieved higher validity and lower cost than the baselines regardless of the datasets.

\section{Additional Comments on Existing Assets}\label{section:assets}
Numba 0.56.4\footnote{\url{https://numba.pydata.org/}} is publicly available under the BSD-2-Clause license. 
All the scripts and datasets used in our experiments are available in our GitHub repository at \url{https://github.com/kelicht/ract}. 

All datasets used in \cref{sec:exp} are publicly available and do not contain any identifiable information or offensive content. 
As they are accompanied by appropriate citations in the main body, see the corresponding references for more details.


\begin{table}[p]
    \centering
        \caption{
        Details of each feature of the FICO dataset~\cite{fico:2018}. 
    }
\vskip 0.1in
    \label{tab:appendix:datasets:fico}
    \begin{tabular}{lcccccc}
    \toprule
        \textbf{Feature} & \textbf{Type} & \textbf{Min} & \textbf{Max} & \textbf{Immutable} & \textbf{Constraint} \\
    \midrule
              ExternalRiskEstimate & Integer &  0.0 &  94.0 &     Yes &        Fix \\
             MSinceOldestTradeOpen & Integer &  0.0 & 803.0 &     Yes &        Fix \\
         MSinceMostRecentTradeOpen & Integer &  0.0 & 383.0 &     Yes &        Fix \\
                    AverageMInFile & Integer &  4.0 & 383.0 &     Yes &        Fix \\
             NumSatisfactoryTrades & Integer &  0.0 &  79.0 &      No &    Nothing \\
       NumTrades60Ever2DerogPubRec & Integer &  0.0 &  19.0 &     Yes &        Fix \\
       NumTrades90Ever2DerogPubRec & Integer &  0.0 &  19.0 &     Yes &        Fix \\
            PercentTradesNeverDelq & Integer &  0.0 & 100.0 &      No &    Nothing \\
              MSinceMostRecentDelq & Integer &  0.0 &  83.0 &      No &    Nothing \\
          MaxDelq2PublicRecLast12M & Integer &  0.0 &   9.0 &      No &    Nothing \\
                       MaxDelqEver & Integer &  2.0 &   8.0 &      No &    Nothing \\
                    NumTotalTrades & Integer &  0.0 & 104.0 &     Yes &        Fix \\
            NumTradesOpeninLast12M & Integer &  0.0 &  19.0 &     Yes &        Fix \\
              PercentInstallTrades & Integer &  0.0 & 100.0 &      No &    Nothing \\
      MSinceMostRecentInqexcl7days & Integer &  0.0 &  24.0 &      No &    Nothing \\
                      NumInqLast6M & Integer &  0.0 &  66.0 &      No &    Nothing \\
             NumInqLast6Mexcl7days & Integer &  0.0 &  66.0 &      No &    Nothing \\
        NetFractionRevolvingBurden & Integer &  0.0 & 232.0 &      No &    Nothing \\
          NetFractionInstallBurden & Integer &  0.0 & 471.0 &      No &    Nothing \\
        NumRevolvingTradesWBalance & Integer &  0.0 &  32.0 &      No &    Nothing \\
          NumInstallTradesWBalance & Integer &  0.0 &  23.0 &      No &    Nothing \\
NumBank2NatlTradesWHighUtilization & Integer &  0.0 &  18.0 &      No &    Nothing \\
             PercentTradesWBalance & Integer &  0.0 & 100.0 &      No &    Nothing \\
    \bottomrule
    \end{tabular}
\end{table}
\begin{table}[p]
    \centering
    \caption{
        Details of each feature of the COMPAS dataset~\cite{compas:2016}. 
    }
\vskip 0.1in
    \begin{tabular}{lcccccc}
    \toprule
        \textbf{Feature} & \textbf{Type} & \textbf{Min} & \textbf{Max} & \textbf{Immutable} & \textbf{Constraint} \\
    \midrule
                      age & Integer & 18.0 & 96.0 &      No & Increasing only \\
            juv\_fel\_count & Integer &  0.0 & 20.0 &      No &         Nothing \\
           juv\_misd\_count & Integer &  0.0 & 13.0 &      No &         Nothing \\
          juv\_other\_count & Integer &  0.0 &  9.0 &      No &         Nothing \\
             priors\_count & Integer &  0.0 & 38.0 &      No &         Nothing \\
    race:African-American &  Binary &  0.0 &  1.0 &     Yes &             Fix \\
               race:Asian &  Binary &  0.0 &  1.0 &     Yes &             Fix \\
           race:Caucasian &  Binary &  0.0 &  1.0 &     Yes &             Fix \\
            race:Hispanic &  Binary &  0.0 &  1.0 &     Yes &             Fix \\
     race:Native American &  Binary &  0.0 &  1.0 &     Yes &             Fix \\
               race:Other &  Binary &  0.0 &  1.0 &     Yes &             Fix \\
        c\_charge\_degree:F &  Binary &  0.0 &  1.0 &      No &         Nothing \\
        c\_charge\_degree:M &  Binary &  0.0 &  1.0 &      No &         Nothing \\
                   gender &  Binary &  0.0 &  1.0 &     Yes &             Fix \\
    \bottomrule
    \end{tabular}
    \label{tab:appendix:datasets:compas}
\end{table}
\begin{table}[p]
    \centering
        \caption{
        Details of each feature of the Credit dataset~\cite{Credit:2009}. 
    }
\vskip 0.1in
    \label{tab:appendix:datasets:credit}
    \begin{tabular}{lcccccc}
    \toprule
        \textbf{Feature} & \textbf{Type} & \textbf{Min} & \textbf{Max} & \textbf{Immutable} & \textbf{Constraint} \\
    \midrule
                              Married &  Binary &  0.0 &     1.0 &     Yes &        Fix \\
                               Single &  Binary &  0.0 &     1.0 &     Yes &        Fix \\
                            Age\_lt\_25 &  Binary &  0.0 &     1.0 &     Yes &        Fix \\
                      Age\_in\_25\_to\_40 &  Binary &  0.0 &     1.0 &     Yes &        Fix \\
                      Age\_in\_40\_to\_59 &  Binary &  0.0 &     1.0 &     Yes &        Fix \\
                           Age\_geq\_60 &  Binary &  0.0 &     1.0 &     Yes &        Fix \\
                       EducationLevel & Integer &  0.0 &     3.0 &      No &    Nothing \\
         MaxBillAmountOverLast6Months & Integer &  0.0 & 50810.0 &      No &    Nothing \\
      MaxPaymentAmountOverLast6Months & Integer &  0.0 & 51430.0 &      No &    Nothing \\
 MonthsWithZeroBalanceOverLast6Months & Integer &  0.0 &     6.0 &      No &    Nothing \\
 MonthsWithLowSpendingOverLast6Months & Integer &  0.0 &     6.0 &      No &    Nothing \\
MonthsWithHighSpendingOverLast6Months & Integer &  0.0 &     6.0 &      No &    Nothing \\
                 MostRecentBillAmount & Integer &  0.0 & 29450.0 &      No &    Nothing \\
              MostRecentPaymentAmount & Integer &  0.0 & 26670.0 &      No &    Nothing \\
                   TotalOverdueCounts & Integer &  0.0 &     3.0 &     Yes &        Fix \\
                   TotalMonthsOverdue & Integer &  0.0 &    36.0 &     Yes &        Fix \\
    \bottomrule
    \end{tabular}
\end{table}
\begin{table}[p]
    \centering
        \caption{
        Details of each feature of the Bail dataset~\cite{Bail:1988}. 
    }
\vskip 0.1in
    \label{tab:appendix:datasets:bail}
    \begin{tabular}{lcccccc}
    \toprule
        \textbf{Feature} & \textbf{Type} & \textbf{Min} & \textbf{Max} & \textbf{Immutable} & \textbf{Constraint} \\
    \midrule
      White &  Binary &  0.0 &   1.0 &     Yes &        Fix \\
      Alchy &  Binary &  0.0 &   1.0 &      No &    Nothing \\
      Junky &  Binary &  0.0 &   1.0 &      No &    Nothing \\
      Super &  Binary &  0.0 &   1.0 &     Yes &        Fix \\
    Married &  Binary &  0.0 &   1.0 &     Yes &        Fix \\
      Felon &  Binary &  0.0 &   1.0 &     Yes &        Fix \\
    Workrel &  Binary &  0.0 &   1.0 &      No &    Nothing \\
     Propty &  Binary &  0.0 &   1.0 &     Yes &        Fix \\
     Person &  Binary &  0.0 &   1.0 &     Yes &        Fix \\
       Male &  Binary &  0.0 &   1.0 &     Yes &        Fix \\
     Priors & Integer &  0.0 &  40.0 &     Yes &        Fix \\
     School & Integer &  1.0 &  19.0 &      No &    Nothing \\
       Rule & Integer &  0.0 &  39.0 &      No &    Nothing \\
        Age & Integer & 15.0 &  72.0 &     Yes &        Fix \\
     Tservd & Integer &  1.0 & 287.0 &     Yes &        Fix \\
     Follow & Integer & 46.0 &  57.0 &     Yes &        Fix \\
    \bottomrule
    \end{tabular}
\end{table}

\begin{table*}[p]
    \caption{
        Experimental results on the average running time [s] of our performance comparison. 
        There is no significant difference between the baselines and RACT on almost all the datasets. 
    }
\vskip 0.1in
    \label{tab:appendix:exp:comp:time}
    \centering
    \begin{tabular}{ccccccc}
    \toprule
        & \multicolumn{3}{c}{\textbf{Classification Tree}} & \multicolumn{3}{c}{\textbf{Random Forest}} \\
        \cmidrule(lr){2-4} \cmidrule(lr){5-7}
        \textbf{Dataset} & \textbf{Vanilla} & \textbf{OAF} & \textbf{RACT} & \textbf{Vanilla} & \textbf{OAF} & \textbf{RACT} \\
    \midrule
        FICO & $0.483 \pm 0.04$ & $0.324 \pm 0.02$ & $0.429 \pm 0.06$ & $42.09 \pm 1.8$ & $35.25 \pm 1.52$ & $48.07 \pm 2.33$ \\
        COMPAS & $0.098 \pm 0.02$ & $0.043 \pm 0.01$ & $0.036 \pm 0.0$ & $2.31 \pm 0.32$ & $1.76 \pm 0.14$ & $2.07 \pm 0.13$ \\
        Credit & $1.827 \pm 0.07$ & $1.37 \pm 0.01$ & $1.587 \pm 0.1$ & $292.53 \pm 21.6$ & $268.41 \pm 18.9$ & $277.40 \pm 25.8$ \\
        Bail & $0.337 \pm 0.02$ & $0.054 \pm 0.01$ & $0.2 \pm 0.05$ & $24.84 \pm 1.93$ & $3.26 \pm 0.35$ & $23.04 \pm 1.64$ \\
    \bottomrule
    \end{tabular}
\end{table*}

\begin{table*}[p]
    \centering
        \caption{
        Average cost of extracted actions (lower is better). 
        We used the MPS~\cite{Ustun:FAT*2019} as a cost function $c$. 
        Our RACT attained lower costs than the baselines regardless of the datasets. 
    }
\vskip 0.1in
    \label{tab:appendix:exp:qual:cost}
    \begin{tabular}{ccccccc}
    \toprule
        & \multicolumn{3}{c}{\textbf{Classification Tree}} & \multicolumn{3}{c}{\textbf{Random Forest}} \\
        \cmidrule(lr){2-4} \cmidrule(lr){5-7}
        \textbf{Dataset} & \textbf{Vanilla} & \textbf{OAF} & \textbf{RACT} & \textbf{Vanilla} & \textbf{OAF} & \textbf{RACT} \\
    \midrule
        FICO & $0.324 \pm 0.11$ & $0.174 \pm 0.02$ & $\bm{0.11 \pm 0.02}$ & $0.447 \pm 0.05$ & $0.407 \pm 0.03$ & $\bm{0.283 \pm 0.01}$ \\
        COMPAS & $0.26 \pm 0.02$ & $0.257 \pm 0.02$ & $\bm{0.184 \pm 0.05}$ & $0.298 \pm 0.02$ & $0.28 \pm 0.01$ & $\bm{0.232 \pm 0.02}$ \\
        Credit & $0.265 \pm 0.03$ & N/A & $\bm{0.217 \pm 0.05}$ & $0.293 \pm 0.02$ & N/A & $\bm{0.166 \pm 0.04}$ \\
        Bail & $0.6 \pm 0.09$ & $0.47 \pm 0.03$ & $\bm{0.221 \pm 0.01}$ & $0.763 \pm 0.03$ & $0.525 \pm 0.04$ & $\bm{0.419 \pm 0.05}$ \\
    \bottomrule
    \end{tabular}
\end{table*}
\begin{table*}[p]
    \centering
        \caption{
        Average plausibility of extracted actions (lower is better). 
        Following \cite{Parmentier:ICML2021}, we measured the ourlier score estimated by isolation forests. 
        There is no significant difference on the plausibility between the baselines and our RACT. 
    }
\vskip 0.1in
    \label{tab:appendix:exp:qual:plausibility}
    \begin{tabular}{ccccccc}
    \toprule
        & \multicolumn{3}{c}{\textbf{Classification Tree}} & \multicolumn{3}{c}{\textbf{Random Forest}} \\
        \cmidrule(lr){2-4} \cmidrule(lr){5-7}
        \textbf{Dataset} & \textbf{Vanilla} & \textbf{OAF} & \textbf{RACT} & \textbf{Vanilla} & \textbf{OAF} & \textbf{RACT} \\
    \midrule
        FICO & $0.481 \pm 0.01$ & $\bm{0.46 \pm 0.0}$ & $0.476 \pm 0.0$ & $0.456 \pm 0.0$ & $0.446 \pm 0.0$ & $\bm{0.437 \pm 0.0}$ \\
        COMPAS & $\bm{0.46 \pm 0.01}$ & $\bm{0.46 \pm 0.01}$ & $0.477 \pm 0.03$ & $\bm{0.44 \pm 0.01}$ & $0.447 \pm 0.01$ & $0.453 \pm 0.01$ \\
        Credit & $0.525 \pm 0.01$ & N/A & $\bm{0.506 \pm 0.01}$ & $0.526 \pm 0.01$ & N/A & $\bm{0.523 \pm 0.01}$ \\
        Bail & $0.511 \pm 0.01$ & $\bm{0.506 \pm 0.01}$ & $0.517 \pm 0.01$ & $\bm{0.504 \pm 0.01}$ & $0.507 \pm 0.0$ & $0.512 \pm 0.01$ \\
    \bottomrule
    \end{tabular}
\end{table*}
\begin{table*}[p]
    \centering
        \caption{
        Average validity of extracted actions under the causal recourse constraint proposed by \cite{Karimi:FAccT2021} (higher is better). 
        Our RACT attained higher validity than the baselines even in the causal recourse setting. 
    }
\vskip 0.1in
    \label{tab:appendix:exp:qual:causality}
    \begin{tabular}{ccccccc}
    \toprule
        & \multicolumn{3}{c}{\textbf{Classification Tree}} & \multicolumn{3}{c}{\textbf{Random Forest}} \\
        \cmidrule(lr){2-4} \cmidrule(lr){5-7}
        \textbf{Dataset} & \textbf{Vanilla} & \textbf{OAF} & \textbf{RACT} & \textbf{Vanilla} & \textbf{OAF} & \textbf{RACT} \\
    \midrule
        FICO & $0.554 \pm 0.15$ & $0.799 \pm 0.05$ & $\bm{0.944 \pm 0.04}$ & $0.316 \pm 0.04$ & $0.502 \pm 0.03$ & $\bm{0.648 \pm 0.03}$ \\
        COMPAS & $0.602 \pm 0.04$ & $0.611 \pm 0.04$ & $\bm{0.838 \pm 0.14}$ & $0.599 \pm 0.02$ & $0.642 \pm 0.04$ & $\bm{0.721 \pm 0.02}$ \\
        Credit & $0.531 \pm 0.05$ & N/A & $\bm{0.672 \pm 0.1}$ & $0.669 \pm 0.05$ & N/A & $\bm{0.964 \pm 0.03}$ \\
        Bail & $0.17 \pm 0.1$ & $0.37 \pm 0.07$ & $\bm{0.726 \pm 0.02}$ & $0.174 \pm 0.04$ & $0.201 \pm 0.04$ & $\bm{0.539 \pm 0.03}$ \\
    \bottomrule
    \end{tabular}
\end{table*}



\begin{table}[p]
    \centering
    \caption{
        Average validity of actions extracted by the MILO-based AR method (higher is better). 
        Our RACT attained higher validity than the baselines regardless of the datasets. 
    }
\vskip 0.1in
    \begin{tabular}{cccc}
    \toprule
        \textbf{Dataset} & \textbf{Vanilla} & \textbf{OAF} & \textbf{RACT} \\
    \midrule
        FICO & $0.234 \pm 0.08$ & $0.132 \pm 0.05$ & $\bm{0.896 \pm 0.05}$ \\
        COMPAS & $0.748 \pm 0.08$ & $0.79 \pm 0.06$ & $\bm{0.938 \pm 0.07}$ \\
        Credit & $0.704 \pm 0.08$ & N/A & $\bm{0.976 \pm 0.02}$ \\
        Bail & $0.076 \pm 0.04$ & $0.25 \pm 0.08$ & $\bm{0.724 \pm 0.07}$ \\
    \bottomrule
    \end{tabular}
    \label{tab:appendix:exp:milo:valid}
\end{table}
\begin{table}[p]
    \centering
    \caption{
        Average cost of actions extracted by the MILO-based AR method (lower is better). 
        We used the MPS~\cite{Ustun:FAT*2019} as a cost function $c$. 
        Our RACT attained lower costs than the baselines regardless of the datasets. 
    }
\vskip 0.1in
    \begin{tabular}{cccc}
    \toprule
        \textbf{Dataset} & \textbf{Vanilla} & \textbf{OAF} & \textbf{RACT} \\
    \midrule
        FICO & $0.429 \pm 0.08$ & $0.553 \pm 0.02$ & $\bm{0.217 \pm 0.04}$ \\
        COMPAS & $0.226 \pm 0.03$ & $0.207 \pm 0.02$ & $\bm{0.124 \pm 0.03}$ \\
        Credit & $0.219 \pm 0.03$ & N/A & $\bm{0.119 \pm 0.01}$ \\
        Bail & $0.73 \pm 0.08$ & $0.493 \pm 0.06$ & $\bm{0.236 \pm 0.03}$ \\
    \bottomrule
    \end{tabular}
    \label{tab:appendix:exp:milo:cost}
\end{table}

\end{document}